\documentclass[conference]{IEEEtran}
\IEEEoverridecommandlockouts

\usepackage{cite}
\usepackage{amsmath,amssymb,amsfonts}
\usepackage{graphicx}
\usepackage[dvipsnames]{xcolor}
\def\BibTeX{{\rm B\kern-.05em{\sc i\kern-.025em b}\kern-.08em T\kern-.1667em\lower.7ex\hbox{E}\kern-.125emX}}  
\usepackage{array}
\usepackage{textcomp}
\usepackage{stfloats}
\usepackage{url}
\usepackage{tabularx}
\usepackage{verbatim}
\usepackage{enumitem}
\usepackage{stfloats}
\usepackage[bottom]{footmisc}
\usepackage{booktabs}
\usepackage{siunitx} % Required for alignment
\usepackage{orcidlink}
\usepackage{algpseudocode}
\usepackage[bottom]{footmisc}
\usepackage{mathtools}
\usepackage{subcaption}
\usepackage[linesnumbered,ruled,vlined]{algorithm2e}
\algrenewcommand{\algorithmiccomment}[1]{\hfill// #1}
\algnewcommand{\algorithmicor}{\textbf{ or }}
\algnewcommand{\OR}{\algorithmicor}
\usepackage{balance}

\begin{document}
\title{AutoRank: MCDA Based Rank Personalization for LoRA-Enabled Distributed Learning}
%\title{Towards Initial Rank Personalization for LoRA-Enabled Distributed Learning}
\author{
    \IEEEauthorblockN{Shuaijun Chen\IEEEauthorrefmark{1}\orcidlink{0009-0001-4944-3406},~\IEEEmembership{Member,~IEEE}, Omid Tavallaie\orcidlink{0000-0002-3367-1236}\IEEEauthorrefmark{2},~\IEEEmembership{Member,~IEEE}, 
    Niousha Nazimi\orcidlink{0000-0003-4085-7044}\IEEEauthorrefmark{1},~\IEEEmembership{Member,~IEEE}, \\
    Xin Chen\IEEEauthorrefmark{3}, Albert Y. Zomaya\orcidlink{0000-0002-3090-1059}\IEEEauthorrefmark{1},~\IEEEmembership{Fellow,~IEEE}}
    \IEEEauthorblockA{\IEEEauthorrefmark{1} School of Computer Science, The University of Sydney, Australia}
    \IEEEauthorblockA{\IEEEauthorrefmark{2}Department of Engineering Science, University of Oxford, United Kingdom}
    \IEEEauthorblockA{\IEEEauthorrefmark{3}School of Mathematics and Statistics, Central South University, China}
}

\maketitle
\begin{abstract}
As data volumes expand rapidly, distributed machine learning has become essential for addressing the growing computational demands of modern AI systems. However, training models in distributed environments is challenging with participants hold skew, Non-Independent-Identically distributed (Non-IID) data. Low-Rank Adaptation (LoRA) offers a promising solution to this problem by personalizing low-rank updates rather than optimizing the entire model, LoRA-enabled distributed learning minimizes computational and maximize personalization for each participant. Enabling more robust and efficient training in distributed learning settings, especially in large-scale, heterogeneous systems. Despite the strengths of current state-of-the-art methods, they often require manual configuration of the initial rank, which is increasingly impractical as the number of participants grows. This manual tuning is not only time-consuming but also prone to suboptimal configurations. To address this limitation, we propose AutoRank, an adaptive rank-setting algorithm inspired by the bias-variance trade-off. AutoRank leverages the MCDA method TOPSIS to dynamically assign local ranks based on the complexity of each participant's data. By evaluating data distribution and complexity through our proposed data complexity metrics, AutoRank provides fine-grained adjustments to the rank of each participant's local LoRA model. This adaptive approach effectively mitigates the challenges of double-imbalanced, non-IID data. Experimental results demonstrate that AutoRank significantly reduces computational overhead, enhances model performance, and accelerates convergence in highly heterogeneous federated learning environments. Through its strong adaptability, AutoRank offers a scalable and flexible solution for distributed machine learning.

\end{abstract}


\section{Introduction}\label{introduction}
In the rapidly evolving field of distributed machine learning, technologies such as Federated Learning \cite{mcmahan2017communication}, Edge Computing \cite{9052677, gao2024federated}, Data Parallelism \cite{li2014scaling, dean2012large}, and Split Learning \cite{vepakomma2018split} have gained significant traction. These approaches aim to offload model training and inference processes to participants or edge nodes, reduce communication overhead, lower computational costs, preserve user privacy \cite{10647104}, and enhance system responsiveness. However, the deployment of these technologies presents several challenges. A notable issue is the disparity in computational capabilities between edge devices \cite{shi2016edge}, such as smartphones and IoT devices, and traditional data centers. Edge devices often face constraints in computational power, storage capacity, and energy availability, making it challenging to deploy and train large, complex machine-learning models. Moreover, data distributed across participants often exhibits non-Independent and Identically Distributed (non-IID) patterns, exacerbating the difficulty of achieving efficient and effective model convergence in distributed systems. To address these challenges, researchers have proposed various model compression and acceleration techniques, such as pruning\cite{han2015learning}, quantizing\cite{krishnamoorthi2018quantizing}, knowledge distillation\cite{hinton2015distilling}, etc. Among them, Low-Rank Adaptation (LoRA)\cite{hu2021lora} stands out as an effective solution. LoRA introduces low-rank adaptation matrices into the weight matrices of pre-trained models, requiring only a small number of parameters to be updated during training. This significantly reduces computational demands and memory consumption, enabling model training on resource-constrained devices.

\begin{figure}[t]
  \centering
    \includegraphics[scale=0.41]{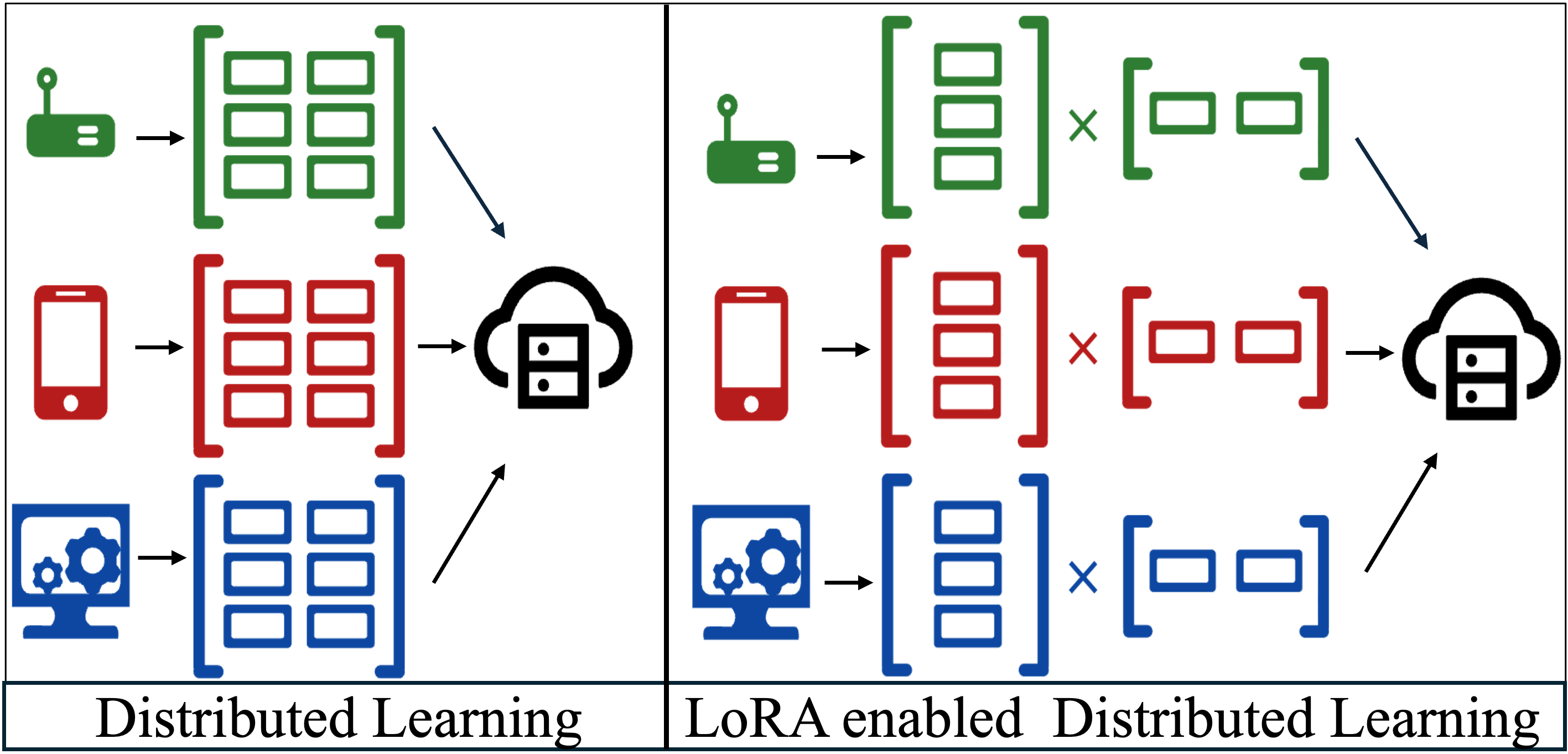}
  \caption{Comparison of standard distributed learning with LoRA-enabled distributed learning. In the LoRA-enabled setup, each participant trains two low-rank models instead of full parameter matrices, reducing communication and computation overhead.}
  \label{fig:lora_enalbed_fl}
\end{figure}

Nevertheless, traditional LoRA methods have limitations. A critical drawback is the need to manually determine the rank of the adaptation matrices. The rank choice directly impacts the model's performance and computational efficiency. A rank that is too low may lead to under-fitting and failing to capture essential data features. Conversely, an excessively high rank can increase computational and storage overhead, undermining the goal of lightweight model design and potentially causing over-fitting, especially when dealing with limited local data. Additionally, non-IID data distributions pose a significant challenge in distributed machine learning scenarios with numerous participants. Participants often have datasets with distinct distributions—for instance, in federated learning, user devices store personalized, locally generated data. Extreme non-IID distributions can slow down the global model's convergence, prolong training time, and degrade overall system performance. LoRA's ability to adaptively adjust the rank of adaptation matrices offers a promising avenue to mitigate these issues. By tailoring the rank to match the data complexity of individual participants, LoRA addresses the computational disparities across edge devices and improves training efficiency in non-IID settings. This paper proposes AutoRank, an automated initial model rank-setting algorithm design for LoRA-enabled distributed machine learning. AutoRank personalizes participant ranks based on their data distribution, aiming to enhance the global model's performance and accelerate convergence while keeping a low computational and communication overhead. The main contributions of this work are summarized as follows:

\begin{enumerate}[topsep=0 pt, partopsep=0 pt, wide=0 pt] \item We analyze the limitations of the commonly used homogeneous local model settings and provide a critical evaluation based on the Bias-Variance Trade-off. Our analysis demonstrates the relationship between global generalization error and local model complexity and the insights behind rank setting.

\item We introduce AutoRank, a flexible and efficient participant rank-setting approach with low computational and communication overhead. And a simple but effective 
multi-prospective data complexity evaluation framework.

\item We implement AutoRank in Python using the TensorFlow library and benchmark its performance against state-of-the-art methods across multiple datasets and configurations. Additionally, we explore the adaptability and robustness of our approach. \end{enumerate}

The remainder of this paper is organized as follows: Section~\ref{relatedwork} discusses related work. Section~\ref{sec:problemstatement} explains the homogeneous model problem and insights behind the rank setting. Section~\ref{sec:contribution} introduces the contribution, including metrics we use to measure data complexity and procedure of AutoRank. Section~\ref{sec:experiment_and_evaluation} discusses the performance between AutoRank and state-of-the-art methods and analyzes the flexibility of AutoRank. Finally, Section~\ref{sec:contribution} concludes this paper.

\section{Related Work}\label{relatedwork}
Distributed machine learning has emerged as a critical area of research, addressing challenges related to scalability, resource constraints, and data privacy. This section explores prior work in key areas relevant to our study: Federated Learning (FL), data parallelism, split learning, incremental learning, Low-Rank Adaptation (LoRA), edge computing, model compression, and model evaluation techniques.

LoRA has emerged as a promising solution for reducing the computational demands of large-scale models. Hu et al. \cite{hu2021lora} introduced LoRA as a method for adapting pre-trained language models with low-rank updates, which reduces memory and computational requirements while maintaining performance comparable to full fine-tuning \cite{howard2018universal}. In federated learning, techniques like LoRA are particularly beneficial for addressing resource heterogeneity among participants. FLoCoRA \cite{grativol2024flocora}, a recent advancement, demonstrates how LoRA can be applied to train small vision models in FL from scratch. Unlike earlier methods that focused on large models, in FLoCoRA only the LoRA adapter parameters are trained, exchanged, and updated, thus the original model is kept frozen. This approach reduces the number of trainable parameters and decreases memory requirements during training. QA-LoRA \cite{xu2023qa} builds on the LoRA framework by introducing quantization-aware adaptation, which combines low-rank adaptation with low-bit quantization to reduce memory requirements during fine-tuning.  Similarly, Dettmers et al. \cite{dettmers2024qlora} presented QLoRA, an efficient fine-tuning approach that utilizes 4-bit quantized models combined with LoRA adapters to reduce memory usage without compromising performance.
FL systems address challenges posed by heterogeneous participant data and resource disparities. McMahan et al. in \cite{mcmahan2017communication} proposed Federated Averaging (FedAvg) aggregate model updates from decentralized data sources efficiently. While FedAvg demonstrated robustness against communication constraints, it implicitly assumes homogeneity in participant capabilities and relies on balanced, Independent and Identically Distributed (IID) datasets for optimal performance. These assumptions limit its applicability in real-world settings characterized by non-IID data and diverse participant behaviors. Several studies have addressed these limitations by enhancing FL's resilience to heterogeneity. Wang et al. \cite{wang2024flora} extended the FL framework by proposing FLoRA, a method designed to support aggregation of LoRA modules with varying ranks in scenarios with heterogeneous participant resources and non-uniform data distributions. FLoRA utilizes a stacking-based noise-free aggregation technique that preserves mathematical accuracy and eliminates errors introduced by naive averaging, as seen in earlier approaches, such as FedIT \cite{zhang2024towards}. Authors in FedIT introduced a framework for instruction tuning of large language models (LLMs), which adapts LoRA modules for parameter-efficient fine-tuning on decentralized participant datasets to improve model generalization across diverse and heterogeneous instruction sets. 

Data parallelism has been widely used to distribute computational workloads across multiple nodes or devices, enabling efficient training of large-scale machine learning models. Li et al. \cite{li2014scaling} introduced the parameter server framework, which addresses scaling distributed machine learning by partitioning data and model parameters across nodes. Similarly, Google in \cite{dean2012large} proposed the DistBelief framework that highlighted the use of data and model parallelism to train large-scale deep neural networks (billions of parameters) across distributed systems. DistBelief introduced methods such as Downpour SGD (asynchronous stochastic gradient descent) and Sandblaster L-BFGS to optimize training efficiency and scalability. Recent advancements in distributed machine learning, building upon parameter server and DistBelief, have introduced innovative frameworks that enhance the efficiency of training large-scale models. For instance, AutoDDL \cite{chen2024autoddl} is a distributed training framework designed to optimize the combination of data, operator, and pipeline parallelism. AutoDDL automates finding near-optimal bandwidth efficiency by employing the Split Broadcast Partial-value (SBP) framework provided in OneFlow \cite{yuan2021oneflow}, which abstracts tensor operations for distributed environments. SBP improves scalability by enabling hybrid parallelism that allows tensors to be split across dimensions, broadcasted, or partially summed. Similarly, the PyTorch Distributed module \cite{li2020pytorch} has been designed to accelerate data-parallel training by implementing techniques such as gradient bucketing, overlapping computation with communication, and skipping gradient synchronization. These methods enable the system to achieve near-linear scalability across multiple GPUs, improving efficiency for large-scale distributed machine-learning tasks.

Split learning has gained prominence as a privacy-preserving technique in distributed systems. Vepakomma et al. \cite{vepakomma2018split} proposed Split Learning (SplitNN) for healthcare applications, where data remains local while intermediate computations are shared between nodes. This approach addresses privacy concerns by ensuring that sensitive patient data is not exchanged among participants.  A recent advancement, CURE \cite{kanpak2024cure}, builds on SplitNN by integrating homomorphic encryption (HE) \cite{gentry2009fully} to enhance privacy in split learning setups. Unlike traditional SplitNN, which may leak sensitive information through intermediate computations, CURE encrypts the server-side model parameters and data; thus, intermediate outputs cannot reveal sensitive participant information.

\section{Problem Statement}\label{sec:problemstatement}
In this section, we introduce the crucial problem of cross-participant homogeneous models in current distributed learning systems under double imbalance data and insights into model complexity personalization.

\subsection{Cross-participant Homogeneous Model}

\begin{figure}[b]
    \centering
    \begin{subfigure}[b]{0.24\textwidth}
        \includegraphics[width=45 mm, height=42.5 mm]{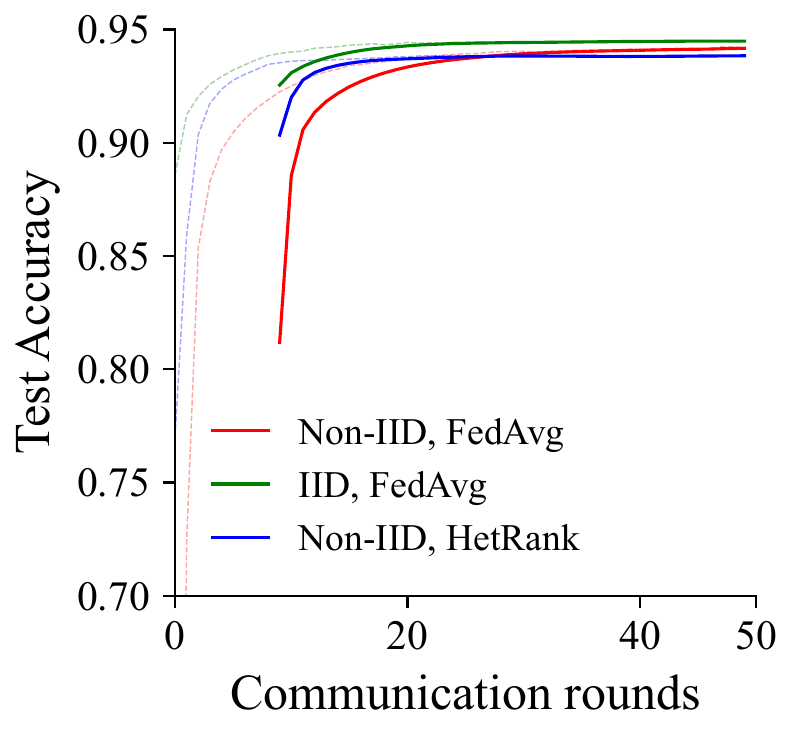}
        \caption{The impact on model convergence under non-IID data.}
        \label{fig:noniid}
    \end{subfigure}
%    \hfill
    \begin{subfigure}[b]{0.24\textwidth}
        \includegraphics[width=45 mm, height=42.5 mm]{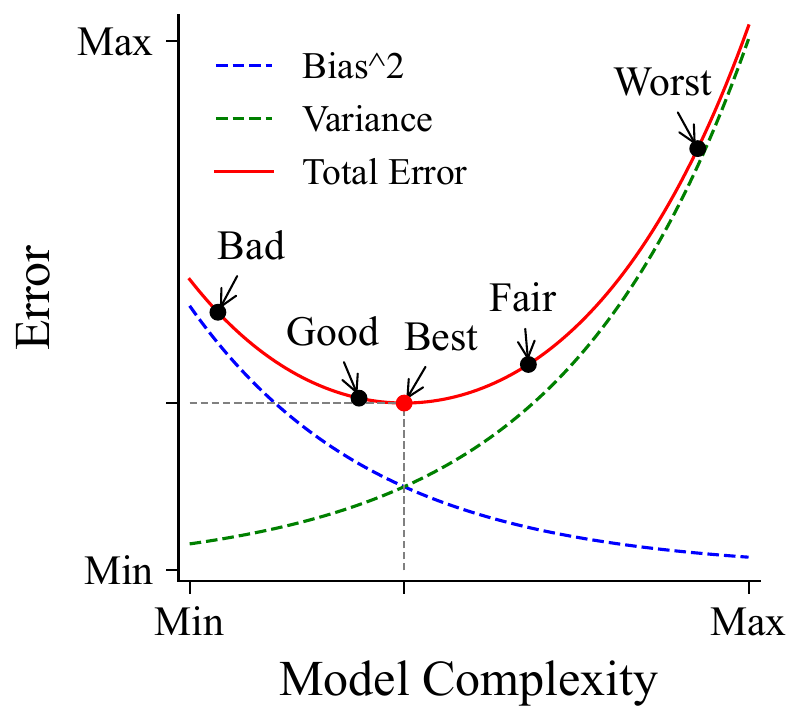}
        \caption{Illustration of Bias-Variance Trade-off.}
        \label{fig:bias_variance_tradeoff}
    \end{subfigure}
    \caption{Fig.~\ref{fig:noniid} compares model convergence under different data distributions. Fig.~\ref{fig:bias_variance_tradeoff} illustrates the bias-variance trade-off, showing how increasing model complexity reduces bias but increases variance and good or bad of different model complexity choices.}
    \label{fig:problem_statement}
\end{figure}

Existing low-rank distributed machine learning methods typically assume a uniform matrix size and rank for all participants, lacking a principled way to initialize those ranks. Under scenarios of "double imbalance"—where \textbf{both data volume and label distribution are skewed}—this uniform approach can produce severe \textbf{over-fitting for some participants and under-fitting for others}, ultimately slowing global model convergence and harming generalization. While the personalized low rank model can partially reduce the over-fitting and under-fitting by adaptively assigning different ranks to participants. Fig.~\ref{fig:noniid} shows the experiments conducted on MNIST comparing Homogeneous-Model FedAvg and a Heterogeneous-Rank LoRA-enabled FL. In the experiment, one participant possesses all labels. Still, each label contains only 1/10 of the original samples, while another participant has data for only a single label but with an equal number of samples as the first client. In the homogeneous setting, both clients utilize the same complete model. However, in the heterogeneous setting, the client with only one label is assigned a model with approximately 1/10 of the trainable parameters of the other. The results show that the homogeneous model setting significantly slows down the convergence rate compared to the heterogeneous setting. Furthermore, the heterogeneous setting \textbf{reduces communication and computation costs by approximately 45\%} while delivering even better performance. Although good performance can be achieved, the current settings rely on manual configuration. As the number of participants increases, manual configuration becomes impractical. Therefore, it is essential to develop an algorithm that can automatically adapt the model's rank for each participant.

%\vspace{-3mm}

\subsection{Insights of rank setting}
We use federated learning as a case for analyzing the local generalization error $Err_i$ for a participant $i$ using squared loss can be decomposed as $Err_i = \sigma_\varepsilon^2 + \left[ \mathbb{E}\hat{f}(x_i) - f(x_i) \right]^2 + \mathbb{E}\left[ \hat{f}(x_i) - \mathbb{E}\hat{f}(x_i) \right]^2$ where $x_i$ represents participant $ i$'s input. Here, $\left[ \mathbb{E}\hat{f}(x_i) - f(x_i) \right]^2$ corresponds to the squared bias ($\text{Bias}_i^2$), $\mathbb{E}\left[ \hat{f}(x_i) - \mathbb{E}\hat{f}(x_i) \right]^2$ corresponds to the variance($\text{Var}_i$), and $\sigma_\varepsilon^2$ represents the irreducible error. In data volume-based FedAvg, the global model $w_g$ is computed as:

\begin{equation}
    w_g = \sum_{i=1}^K \frac{p_i}{\sum_{j=1}^K p_j} w_i.
\end{equation}

% We denote $p_i$ as the data volume of client $i$ and $w_i$ as the model weights of client $i$. 
% Since the global aggregation process does not involve additional training, the generalization error of the global model $Err_g$, can be expressed by considering each client's bias and variance contributions. 
% Let $\mathcal{C}_i$ represent the model complexity and $\mathcal{D}_i$ denote the data complexity of client $i$. 
% Then we define $\text{Bias}_i=\text{Bias}(\mathcal{C}_i,\mathcal{D}_i)$ and $\text{Var}_i=\text{Var}(\mathcal{C}_i,\mathcal{D}_i)$ as differentiable functions of model and data complexity. 
% Hence, the generalization error is formulated as:

% \begin{align}
%     Err_g = &\underbrace{\sum_{i=1}^K \frac{p_i}{\sum_{j=1}^K p_j} \cdot \text{Bias}^2(\mathcal{C}_i,\mathcal{D}_i)}_{\text{Global Bias}} \;+\;\nonumber \\
%             &\underbrace{\sum_{i=1}^K \frac{p_i}{\sum_{j=1}^K p_j} \cdot \text{Var}(\mathcal{C}_i,\mathcal{D}_i)}_{\text{Global Variance}} \;+\; \sigma_\varepsilon^2.
% \end{align}

where $p_i$ and $p_j$ is the data volume of participant $i$ and participant $j$, and $w_i$ represents the model weights of participant $i$. Since the global aggregation process does not involve any training, the generalization error of the global model, $Err_g$, can be expressed as:

% define the d(C_i)

% \begin{align}
%     \frac{C_i}{d(D_i)} = \frac{\partial \mathcal{C}_i}{\partial \mathcal{D}_i} = - \frac{\partial Err_i}{\partial \mathcal{D}_i} / \frac{\partial Err_i}{\partial \mathcal{C}_i}
% \end{align}

\begin{align}
    Err_g &= \underbrace{\sum_{i=1}^K (\frac{p_i}{\sum_{j=1}^K p_j} \cdot \text{Bias}_i^2}_{\text{Global Bias}} + \underbrace{\frac{p_i}{\sum_{j=1}^K p_j} \cdot \text{Var}_i}_{\text{Global Variance}} + \sigma_\varepsilon^2).\nonumber \\
& = \sum_{i=1}^K \frac{p_i}{\sum_{j=1}^K p_j} \cdot Err_i
\end{align}

We denote $\mathcal{C}_i$ as the model complexity and $\mathcal{D}_i$ as the data complexity (including distribution skewness and data volume) of participant $i$. Define bias and variance as differentiable functions of $\mathcal{C}$ and $\mathcal{D}$ with $\text{Bias}_i=\text{Bias}(\mathcal{C}_i,\mathcal{D}_i),\text{Var}_i=\text{Var}(\mathcal{C}_i,\mathcal{D}_i)$.

We first begin by considering the local error \( {Err}_i \). If we fix local generalization error $Err_i$ as a constant $E_i$ , the relationship between model complexity $\mathcal{C}_i$ and $\mathcal{D}_i$ can be expressed by the following equation:
\begin{align}
    Err_i = \text{Bias}^2(\mathcal{C}_i,\mathcal{D}_i) + \text{Var}(\mathcal{C}_i,\mathcal{D}_i) +  \sigma_\varepsilon^2 = E_i
    \label{alg:err_i}
\end{align}

To establish the relationship between model complexity \( C_i \) and data complexity \( D_i \), we take the total derivative \cite{wainwright2005fundamental} of Eq.~\ref{alg:err_i}:
\begin{align}
    \textit{d} Err_i = \frac{\partial Err_i}{\partial \mathcal{C}_i} \cdot \textit{d}\mathcal{C}_i + \frac{\partial Err_i}{\partial \mathcal{D}_i} \cdot \textit{d}\mathcal{D}_i = 0
\end{align}

Moreover, we can get:
\begin{align}
    \frac{\textit{d} \mathcal{C}_i}{\textit{d} \mathcal{D}_i} = - \frac{\partial Err_i}{\partial \mathcal{D}_i} / \frac{\partial Err_i}{\partial \mathcal{C}_i}
\end{align}

\textcolor{black}{For a fixed $\mathcal{D}_i, $ the plots of the functions \(\text{Bias}(\mathcal{C}_i) \) and \( \text{Var}(\mathcal{C}_i) \) can be found in Fig.~\ref{fig:bias_variance_tradeoff}. Then $\frac{\partial Err_i}{\partial \mathcal{C}_i} < 0$ implies that $\mathcal{C}_i$ is on the left of the extreme point (see Fig.~\ref{fig:bias_variance_tradeoff}), at this point, an increase in \(\mathcal{ D}_i \) will lead to an increase in \( Err_i \), thus $\frac{\partial Err_i}{\partial \mathcal{D}_i}>0,\frac{\partial \mathcal{C}_i}{\partial \mathcal{D}_i}>0$. On the other hand,  $\frac{\partial Err_i}{\partial \mathcal{C}_i} > 0$ implies that $\mathcal{C}_i$ is on the right of the extreme point, thus an increase in \(\mathcal{ D}_i \) will lead to a decrease in \( Err_i \), thus $\frac{\partial Err_i}{\partial \mathcal{D}_i}<0,\frac{\partial \mathcal{C}_i}{\partial \mathcal{D}_i}>0$. We get $\frac{\text{d}\mathcal{C}_i}{\text{d}\mathcal{D}_i} > 0$ if $\frac{\partial Err_i}{\partial \mathcal{C}_i } \neq 0$. Therefore, under the condition of equal error, the model complexity $\mathcal{C}_i$ is positively correlated with the data complexity $\mathcal{D}_i$  which has already been mentioned in the literature\cite{vapnik2015uniform}\cite{hastie2009elements}, can be expressed as:}
\begin{equation}
    \mathcal{C}_i \propto \mathcal{D}_i 
    \quad \text{(for fixed \(\mathrm{Err}_i\))}.
\end{equation}

This result demonstrates that as data complexity $\mathcal{D}_i$ increases, the model complexity $\mathcal{C}_i$ must also increase to maintain a constant error level. 
%  Thus, Equation (9) can be transformed into the following form:
% \begin{align}
%     Err_g = \sum_{i=1}^K \frac{p_i}{\sum_{j=1}^K p_j} \cdot \text{Bias}^2(\mathcal{C}_i,\mathcal{D}_i) + \nonumber \\
%     \sum_{i=1}^K \frac{p_i}{\sum_{j=1}^K p_j} \cdot \text{Var}(\mathcal{C}_i,\mathcal{D}_i) + \sigma_\varepsilon^2.
% \end{align}
Building on the relationship between model complexity $\mathcal{C}_i$ and data complexity $\mathcal{D}_i$, we now analyze how $\mathcal{C}_i$ influences the global generalization error $Err_g$. Assuming that the $\text{Bias}(\mathcal{C},\mathcal{D})$ and $\text{Var}(\mathcal{C},\mathcal{D})$ functions are differentiable, the derivative of $Err_g$ with respect to $\mathcal{C}_i$ can be expressed as:

\begin{align}
    \frac{\partial{Err_g}}{\partial{\mathcal{C}_i}} &= \frac{p_i}{\sum_{j=1}^{K}p_j} \cdot [\frac{\partial{\text{Bias}^2(\mathcal{C}_i,\mathcal{D}_i)}}{\partial {\mathcal{C}_i}}+\frac{\partial{\text{Var}(\mathcal{C}_i,\mathcal{D}_i)}}{\partial{\mathcal{C}_i}}]\nonumber \\
    &=\frac{p_i}{\sum_{j=1}^{K}p_j} \cdot \frac{\partial Err_i}{\partial C_i}
\end{align}
 Here, $p_i$ represents the data volume of participant $i$, and $\frac{\partial \text{Bias}^2(C_i, D_i)}{\partial \mathcal{C}_i}$ and $\frac{\partial \text{Var}(C_i, D_i)}{\partial \mathcal{C}_i}$ quantify the sensitivity of the bias and variance terms to changes in model complexity $C_i$. For participant $i$, if we fix \( D_i \), by Fig. 2(b), we define: $\mathcal{C}_{\mathcal{D}_i} = \text{argmin}_{\mathcal{C}_i} Err_i$ It is clear that : 
 \begin{align}
      \frac{\partial Err_g}{\partial \mathcal{C}_i} 
        \begin{cases} 
        < 0, & \text{if } \mathcal{C}_i < \mathcal{C}_{\mathcal{D}_i}, \\ 
        > 0, & \text{if } \mathcal{C}_i > \mathcal{C}_{\mathcal{D}_i}.
        \end{cases}
 \end{align} 

\textcolor{black}{Thus an increase in \(\mathcal{ C}_i \) leads to a decrease in bias and $\frac{\partial{\text{Bias}^2(\mathcal{C}_i,\mathcal{D}_i)}}{\partial {\mathcal{C}_i}}$ , as well as an increase in variance. The local generalization error $Err_i$ also decreases as $\mathcal{C}_i$ increases until $\mathcal{C}_i = \mathcal{C}_{\mathcal{D}_i}$, after which it gradually increases with further increases in $\mathcal{ C}_i$, therefore the global generalization error $Err_g$, like $Err_i$, will first decrease and then increase as $\mathcal{C}_i$ increases.} In the real world scenario, $D_i$ is fixed for each participant. Therefore, for high-complexity data, we should use high-complexity models for training. However, the model complexity should not be excessively high, as this would increase the variance and thus lead to higher error. For participant \( i \), keeping \(\mathcal{C}_i \) close to \( \mathcal{C}_{\mathcal{D}_i} \) can significantly reduce the local error $Err_i$, thereby reducing the global error $Err_g$ as well .

% If the goal is to minimize error, models with varying complexities should be allocated to datasets with different levels of complexity. Specifically, if the model complexity \( \mathcal{C}_i \) initially assigned to client \( i \) is too low to adequately capture the complexity \(\mathcal{ D}_i \) of the dataset, this will result in a large \( \text{Bias}^2(\mathcal{C}_i) \). In such cases, increasing the model complexity \(\mathcal{C}_i \) is necessary to reduce \( \text{Bias}^2(\mathcal{C}_i) \), thereby decreasing the error \( Err_i \) and $Err_g$. Conversely, if the initial model complexity is excessively high, it will lead to a large \( \text{Var}(\mathcal{C}_i) \). In this scenario, reducing the model complexity \(\mathcal{C}_i \) is required to lower \( \text{Var}(\mathcal{C}_i) \), thus also reducing the error

% In summary, assigning models with complexities that match the complexity of the data is crucial for training, as this approach minimizes the generalization error of the global model \( {Err}_g \).

In our framework, we use the participant's LoRA rank $R_i$ to control $C_i$, A key step in this process is determining the rank $R_i$ for each participant. Within our framework, we first set a global LoRA model rank $R_g$, which is designed to ensure that the global LoRA model maintains a comparable number of trainable parameters to the original model. Each participant's rank $R_i$ is then calculated using the $R_i = R_g \cdot r_i$, where $r_i$ represents the rank scale ratio specific to participant $i$. Based on analysis, improper scaling can lead to either under-fitting or over-fitting of the model for certain participants, ultimately impacting the performance of the global training process. Thus, designing a robust and effective method for setting $r_i$ is a key issue and the primary focus of this work. These insights demonstrate the need for a solution that adapts model ranks based on the varying data complexities across participants. To address this challenge, we propose AutoRank, a method that dynamically adjusts local model ranks using fine-grained metrics for data complexity estimation, achieving a balance between computational efficiency and model performance.

% The challenge lies in determining appropriate values for $r_i$ that effectively balance the model's capacity with the client's data complexity. Since $r_i$ directly controls the rank $R_i$, it plays a critical role in adapting the global LoRA model to each client's specific requirements. Improper scaling can lead to either underutilization or overfitting of the model for certain clients, ultimately impacting the performance of the global training process. Thus, designing a robust and effective method for setting $r_i$ is a key issue and the primary focus of this work.

\section{Contribution}\label{sec:contribution}

We have analyzed that participants with more complex data require higher model complexity to more effectively and rapidly reduce the global model's generalization error. Since the true value of $D_i$ is not defined in the bias-variance trade-off, we defined three metrics to evaluate participant data complexity from the perspectives of: dataset learning difficulty, data volume and distribution uniformity. Dataset learning difficulty reflects how challenging it is for a model to fit the data and distribution uniformity considers the heterogeneity of data distributions across participants. This section presents the three metrics we developed to quantify participant's data complexity. And the procedure of TOPSIS method used to evaluate our proposed metrics. 

\subsection{Data complexity estimation}
% To comprehensively assess the complexity of a client's training data, we consider multiple criteria, including the dataset learning difficulty, the uniformity of data distribution, and the data volume.

\subsubsection{Dataset learning difficulty}

\begin{figure}[t]
    \centering
    \begin{subfigure}[b]{0.24\textwidth}
        \includegraphics[height=42.5 mm]{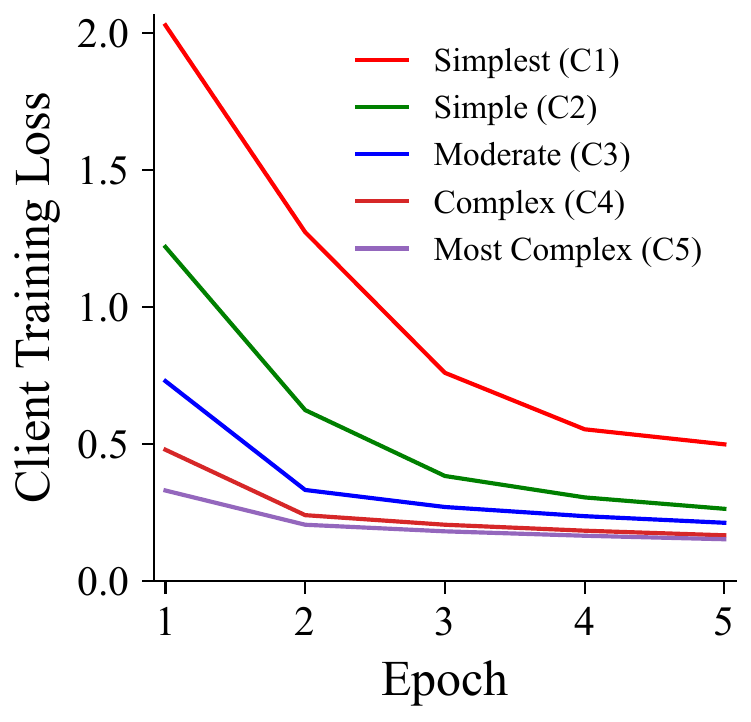}
        \caption{Training loss by epoch under different data complexity.}
        \label{fig:loss_trend}
    \end{subfigure}
    \begin{subfigure}[b]{0.24\textwidth}
        \includegraphics[height=42.5 mm]{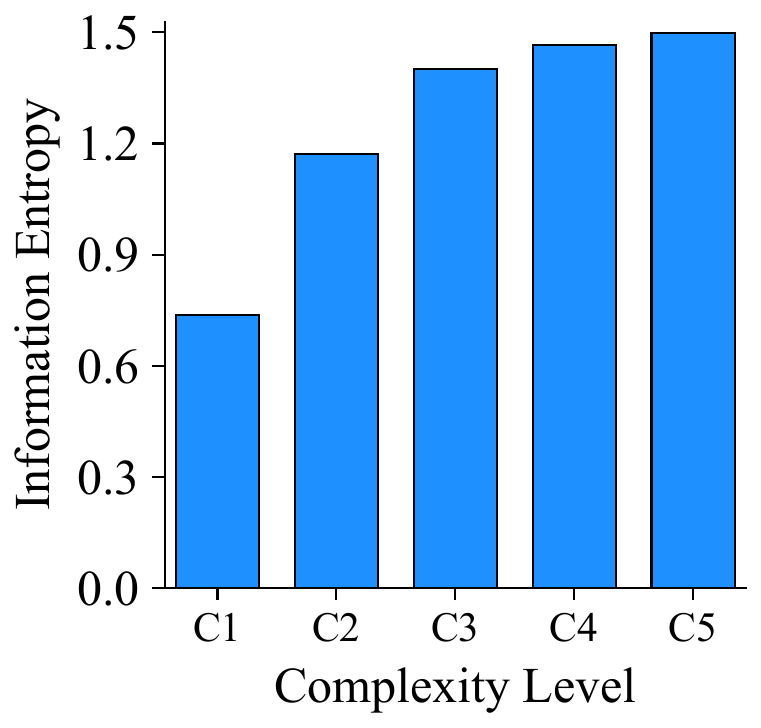}
        \caption{Loss information entropy comparison of Fig.~\ref{fig:loss_trend}.}
        \label{fig:loss_information_entropy_comparison}
    \end{subfigure}
    \caption{Relationship between complexity levels and learning difficulty metric. The left panel shows participant training loss across 5 epochs for different complexity levels (C1 to C5, from simplest to most complex). The right panel displays the information entropy corresponding to each complexity level, highlighting the increasing complexity from C1 to C5.}%\vspace{-4 mm}
    \label{fig:loss_information_entropy}
\end{figure}

The objective is to assign higher ranks to participants with larger and more complex datasets. When the data distribution is uniform, each category's samples contribute equally to the decision boundary, requiring the model to simultaneously focus on multiple categories. This dynamic increases data complexity. Participants with larger datasets and well-balanced distributions contribute more effectively to global model training by providing diverse and representative samples. Assigning appropriate ranks to such participants enhances their data utility while ensuring a balanced and efficient global training process. Fig.~\ref{fig:loss_trend} illustrates the significant differences in training loss trends across participants with varying data complexities. In the figure, data complexity increases progressively from configuration C1 to C5, encompassing both data volume and diversity (number of classes). The results reveal that \textbf{participants with more complex data, such as balanced or diverse distributions, exhibit higher variability and slower convergence in their loss curves}. This observation suggests that the complexity of a participant's data can be effectively quantified using the information entropy of training loss across epochs. In contrast, participants with simpler data distributions achieve faster loss convergence and lower variability, reflecting lower data complexity. These trends emphasize the utility of leveraging the information entropy of training loss as an indirect measure of data complexity in distributed learning scenarios.

%, this kind of data distribution is particularly common in IoT scenarios, where devices with varying capabilities generate data in uneven amounts. For example, low-powered devices like simple sensors or wearables capture limited data, resulting in smaller datasets, while more advanced devices, such as cameras or smart appliances, produce larger and more diverse datasets.

\begin{equation}
    H(L)_i = {-\sum_{e \in E}p_i^{e}*ln(p_i^{e})}, \;\;\; p_i^{e} = \frac{\bar{l}_i^{e}}{ \sum_{e \in E}\bar{l}_i^{e}.}
    \label{information_entropy_loss}
\end{equation}

Based on this observation, we calculate each participant's information entropy $H(L)_i$ of the training loss sequence using Eq.~\ref{information_entropy_loss}, $\bar{l}_i^{e}$ represents the average batch loss for epoch $e$ of participant $i$. This metric estimates the complexity of information contained in a participant's dataset. Participants with more labels and larger datasets generally provide smoother gradient updates during training~\cite{shwartz2017opening}. The increased diversity and volume of data help to average out noise and reduce gradient fluctuations, leading to more stable and consistent updates. This stands in contrast to over-fitted models trained on smaller, less diverse datasets~\cite{neyshabur2017exploring}. Consequently, models trained on larger and more diverse datasets exhibit faster convergence and better generalization performance.

\subsubsection{Distribution estimation}

To perform a fine-grained observation to give a precise rank setting, we introduce a custom entropy and the Gini-Simpson Index to capture adjective differences in the label distribution across participant data. Eq. \ref{information_entropy_label} defines the custom entropy method used to quantify the label distribution, where $p_i^l$ is the probability of label $l$, and $n_i^l$ is the number of samples for label $l$:

\begin{equation} 
H(Y)_i = -\ln\left(\sum_{l \in L} n_i^{l}\right) \sum_{l \in L} p_i^{l} \ln(p_i^{l}),\;\; p_i^{l} = \frac{n_i^{l}}{\sum_{l \in L} n_i^{l}}
\label{information_entropy_label} 
\end{equation}

The custom entropy method specifically considers both the uniformity of the distribution and the data quantity. Traditional information entropy primarily measures the overall uncertainty within a distribution but does not account for the sample size. Since the data quantity significantly impacts the aggregation process, we incorporate the logarithm of the total sample size into the classic entropy calculation, as shown in Equation~\ref{information_entropy_label}. This metric is designed to reflect a combined value of distribution uniformity and data quantity. It ensures that the complexity of the distribution is appropriately weighted based on the data quantity. Given that a larger sample size is not always better, we use a logarithmic function to control the impact of data quantity on the weighting. Moreover, the inclusion of a sample size adjustment factor, $\ln\left(\sum n_i^l\right)$, ensures that the impact of a participant's biased data distribution does not disproportionately dominate the entropy value. Without this adjustment, due to the similar data distribution, entities such as large retail stores and convenience shops might receive misleadingly similar scores despite substantial differences in number of their data labels. For participants with smaller data volumes but highly uneven label distributions, rare labels are prominently reflected in the entropy. However, the adjustment factor mitigates overemphasis on small datasets. Conversely, for participants with larger data volumes, the adjustment factor reduces their excessive influence on the overall entropy value, achieving a balanced representation of label distributions.

\begin{equation} G_i = 1 - \sum_{l \in L} (p_i^l)^2, \label{gini_simpson} \end{equation}

Since information entropy is highly sensitive to low-probability events, participant data distributions in practical non-IID scenarios often exhibit imbalances, with certain labels occurring at very low probabilities. For example, consider two similar distributions (0.5, 0.5, 0) and (0.5, 0.49, 0.01), where the information entropy of the first distribution is 1, while that of the second is 1.07. In such cases, rare labels can disproportionately influence the entropy result, making it less reflective of the overall distribution. To mitigate this issue and enable more meaningful comparisons, we introduce the Gini-Simpson Index, defined in Eq. \ref{gini_simpson}, to reduce the impact of low-probability labels. Unlike entropy, which is logarithmically sensitive and can be heavily influenced by rare events, the Gini-Simpson Index offers a more stable measure by balancing the contributions of all labels, particularly those with low occurrence. As a result, the Gini-Simpson Index provides a more robust and reliable assessment of the data distribution.

% Uniformity directly impacts the model's generalization ability. When a client's data distribution is relatively uniform, it encompasses a broader range of features and patterns. In such scenarios, the local model must exhibit a higher generalization capacity to capture these diverse features, thereby warranting a higher rank assignment. This prevents underfitting and ensures the model fully learns and represents the complex patterns inherent in the data.

% Skewed data distributions can severely impact the generalization of the global model, leading to over-fitting on the data of specific clients. To mitigate the influence of skewed data distributions on model generalization, we assign lower ranks to clients with highly imbalanced data. This reduces the model's complexity, preventing it from over-fitting to the idiosyncratic features of individual clients' data.

\subsection{Evaluation under TOPSIS}
In real-world scenarios, pinpointing a single optimal solution is highly challenging due to the complexity and heterogeneity of data and objectives. Consequently, our algorithmic design is naturally aligned with the principles of Multi-Criteria Decision Analysis (MCDA), which emphasizes a balanced examination of multiple performance metrics. TOPSIS (Technique for Order of Preference by Similarity to Ideal Solution)\cite{tzeng2011multiple} stands out in this regard for several reasons. First, its \textbf{high flexibility allows complete customization}, providing ample trade-off space to accommodate evolving or even conflicting priorities in heterogeneous environments. Second, TOPSIS maintains \textbf{linear computational complexity} ($O(|C||M|)$, where $|C|$ and $|M|$ represent number of participant and number of metrics, respectively), making it suitable for large-scale or real-time applications without overwhelming system resources. Finally, \textbf{by effectively combining multiple distinct metrics} to pinpoint alternatives closest to the ideal solution and farthest from the negative ideal solution, TOPSIS seamlessly integrates with our multi-metric evaluation process, ensuring that the ranking outcomes directly inform and refine our overall algorithmic strategy. Leveraging these strengths, we employ TOPSIS as our primary evaluation framework to comprehensively assess participant-side data complexity. This method not only delivers transparent and interpretable rankings but also provides a robust platform for iterative refinements in rank settings, thus optimally aligning with our system's multi-faceted objectives and real-world application demands.

\subsubsection{Constructing the Weighted Normalized Decision Matrix}
To eliminate the influence of different units and scales, we construct a weighted normalized decision matrix. The normalized value $\bar{x}_{i}^{k}$ and weighted values $v_{i}^{k}$ for each participant $i$ and metric $k$ are calculated as:
\begin{equation}
    \bar{x}_{i}^{k} = \frac{x_{i}^{k}}{\sqrt{\sum_{i \in C} (x_{i}^{k})^2}}, \quad v_{i}^{k} = w^{k} \cdot \bar{x}_{i}^{k}, 
    \label{normalization}
\end{equation}
where $x_{i}^{k}$ is the original value of metric $k$ for participant $i$, $C$ is the set of all participants, and $w^{k}$ represents the weight of metric $k$. The weights $w^{k}$ are determined using the CRITIC method introduced in Section~\ref{section:CRITIC}. The normalized values $\bar{x}_{i}^{k}$ eliminate differences in units and scales across metrics, while the weighted values $v_{i}^{k}$ reflect the relative importance of each metric, forming the final weighted normalized decision matrix.

\subsubsection{Identifying the Ideal and Negative-Ideal Solutions}
The ideal solution (Highest of the metric) $v^{k+}$ and the negative-ideal solution (Lowest of the metric) $v^{k-}$ are defined based on the weighted normalized values:
\begin{equation}\small
    v^{k+} = \left\{ \max_{i} v_{i}^{k} \;|\; k \in M \right\}, \;v^{k-} = \left\{ \min_{i} v_{i}^{k} \;|\; k \in M \right\}
    \label{ideal_negative_solution}
\end{equation}
% \begin{equation}
%     A^- = \left\{ \min_{i} v_{i}^{k} \;|\; k \in M \right\},
%     \label{negative_ideal_solution}
% \end{equation}
where $M$ is the set of all metrics.

\subsubsection{Calculating the Separation Measures}
The separation of each participant from the ideal and negative-ideal solutions is calculated using Euclidean distance:
\begin{equation}
    S_{i}^{+} = \sqrt{\sum_{k \in M} \left( v_{i}^{k} - v^{k+} \right)^2}, \;S_{i}^{-} = \sqrt{\sum_{k \in M} \left( v_{i}^{k} - v^{k-} \right)^2}
    \label{separation_ideal}
\end{equation}
% \begin{equation}
%     S_{i}^{-} = \sqrt{\sum_{k \in M} \left( v_{i}^{k} - v^{k-} \right)^2},
%     \label{separation_negative_ideal}
% \end{equation}
where $v^{k+}$ and $v^{k-}$ are the elements of $A^+$ and $A^-$, respectively.

\subsubsection{Calculating the Relative Closeness to the Ideal Solution}
The relative closeness of each participant to the ideal solution is computed as:
\begin{equation}
    C_{i} = \frac{S_{i}^{-}}{S_{i}^{+} + S_{i}^{-}}.
    \label{relative_closeness}
\end{equation}
A higher value of $C_{i}$ indicates that participant $i$ is closer to the maximum data complexity among all participants. Therefore, participants with higher data complexity will be assigned a higher rank.

\subsection{Weights Calculation using the CRITIC Method}\label{section:CRITIC}

To effectively evaluate multiple metrics within a MCDA framework, it's crucial to assign weights to each metric based on their importance and independence. We employ the CRITIC (Criteria Importance Through Inter-criteria Correlation) method \cite{10.1016/0305-0548(94)00059-H} to calculate these weights. By considering both the contrast intensity (standard deviation) and the correlation (conflict) among criteria, CRITIC systematically balances and integrates potentially conflicting or overlapping metrics, ensuring an objective final evaluation.

\subsubsection{Calculating Standard Deviation}
First, we calculate the standard deviation for each criterion $k$ to measure the variability of the data:

\begin{equation}
\sigma^{k} = \sqrt{\frac{1}{N} \sum_{i=1}^{N} \left( x_{i}^{k} - \bar{x}^{k} \right )^{2}},
\label{standard_deviation}
\end{equation}

where $x_{i}^{k}$ is the value of criterion $k$ for participant $i$, $\bar{x}^{k}$ is the mean value of criterion $k$, and $N$ is the number of participants.

\subsubsection{Calculating the Correlation Coefficient Matrix}
Next, we compute the correlation coefficient matrix $R$, where each element $r^{k j}$ represents the correlation coefficient of participant $i$ between it's criteria $k$ and $j$:

\begin{equation}
r^{k, j} = \frac{\sum_{i=1}^{N} \left( x_{i}^{k} - \bar{x}^{k} \right ) \left( x_{i}^{j} - \bar{x}^{j} \right )}{\sqrt{\sum_{i=1}^{N} \left( x_{i}^{k} - \bar{x}^{k} \right )^{2} \sum_{i=1}^{N} \left( x_{i}^{j} - \bar{x}^{j} \right )^{2}}},
\label{correlation_coefficient}
\end{equation}

for all $k, j \in M$, where $M$ is the set of all criteria.

\subsubsection{Calculating the Amount of Information}
The amount of information $I_{k}$ for each criterion $k$ is calculated by combining the standard deviation and the correlation coefficients:

\begin{equation}
I_{k} = \sigma^{k} \times \left( 1 - \frac{1}{n-1} \sum_{\substack{j=1 \\ j \neq k}}^{n} r^{k, j} \right),
\label{information_content}
\end{equation}

Where $n$ is the total number of criteria. The term $\left( 1 - \frac{1}{n-1} \sum_{j \neq k} r^{k, j} \right)$ represents the average conflict between criterion $k$ and all other criteria.

\subsubsection{Calculating the Weights}
Finally, the weights $w^{k}$ for each criterion $k$ are determined by normalizing the amount of information:

\begin{equation}
w^{k} = \frac{I^{k}}{\sum_{k=1}^{n} I^{k}},
\label{critic_weights}
\end{equation}

ensuring that $\sum_{k=1}^{n} w^{k} = 1$. This normalization ensures that the weights are proportional to the amount of information each criterion contributes.

\begin{figure}[t]
  \centering
    \includegraphics[height = 42.5 mm]{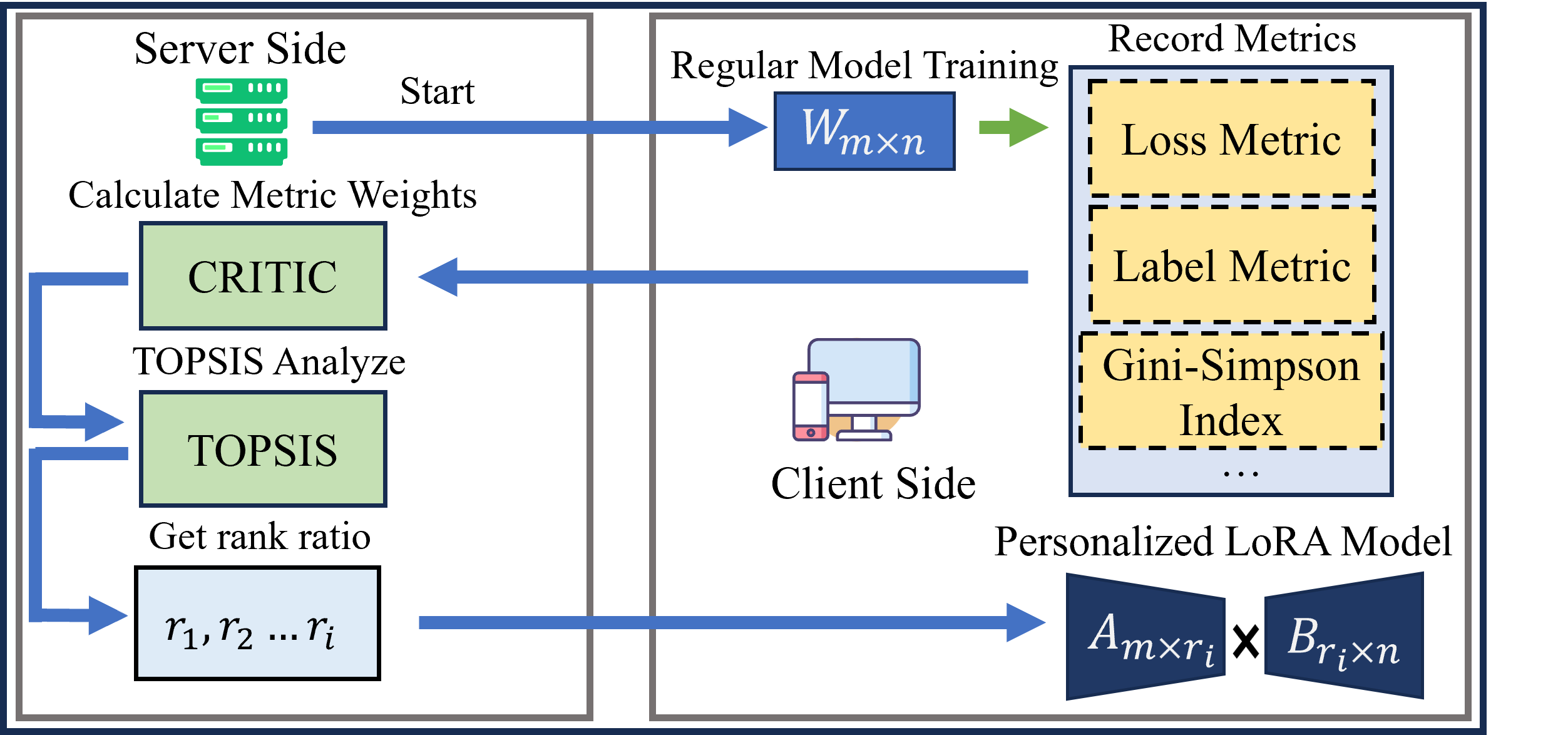}
  \caption{Process of AutoRank.}\vspace{-5mm}
  \label{fig:entire_process}
\end{figure}

\subsection{Put all together}
After obtaining the $C_i$ of each participant, we apply a min-max normalization to scale the values into a standardized range using  Eq. \ref{zero_one_normalize_rank} to calculate the rank scale ratio.

\begin{equation}
    r_i = \max\left(\frac{C_i - C_{\min}}{C_{\max} - C_{\min}},\ \rho \right).
    \label{zero_one_normalize_rank}
\end{equation}

In this context, the normalization process bounds each participant's final rank ratio $r_i$ between a minimum threshold and the maximum normalized value. The whole process of AutoRank is shown in Fig.~\ref{fig:entire_process}, the procedure execute on server side and participant side are concluded in Algorithm.~\ref{alg:rank_setting} and Algorithm.~\ref{alg:client}, respectively.

\subsection{Flexibility of AutoRank}

The versatility of AutoRank largely stems from its foundational evaluation framework, where TOPSIS is employed to flexibly integrate a variety of performance metrics tailored to specific task requirements. This design does not necessitate a complete overhaul of the entire evaluation system whenever new objectives or constraints arise; rather, \textbf{the metrics and their associated weights can be dynamically adjusted} to reflect different priorities. For example, in tasks without explicit labels, one may combine the loss metric with the data volume metric—or even synthesize a custom metric—without losing the overall coherence of the method. Beyond simply accommodating both labeled and unlabeled tasks, AutoRank enables extensive customization through MCDA (TOPSIS) itself. By refining or redefining the metric set and their respective weights, users can shift the focus of the evaluation outcomes to better align with unique goals. For instance, incorporating hardware performance metrics could help assess device capabilities under diverse deployment constraints. In this way, AutoRank offers not just a singular "best" evaluation measure, but rather a flexible foundation that can be readily tuned and extended to fit evolving scenarios. This adaptability facilitates more targeted decision-making, ensuring that AutoRank remains robust and relevant across a wide range of use cases.

\begin{algorithm}[t]
\caption{Procedure of AutoRank exectues on server side.}
\label{alg:rank_setting}

\KwIn{\\
Participants, \( \text{ID} = \{1, 2, \ldots, i\} \),\\
Metrics of each participants: $H(L)_i, H(Y)_i, G_i$}
\KwOut{\\
Rank ratios for each participant: \( \{r_1, r_2, \ldots, r_i\} \)}

Receive metrics data from participants. \\
Apply CRITIC method to compute weights \( \{w_1, w_2, \ldots, w_k\} \) for metrics\;

Use TOPSIS method with weights \( \{w_1, w_2, \ldots, w_k\} \) to calculate the TOPSIS score \( C_i \) for each participant $i$ \;

Perform min-max normalization on \( \{C_1, C_2, \ldots, C_i\} \) to obtain normalized rank ratios \( \{r_1, r_2, \ldots, r_i\} \):
\[
r_i = \max\left(\frac{C_i - C_{\min}}{C_{\max} - C_{\min}},\ \rho \right), \quad \forall i \in \{1, 2, \ldots, N\}
\]
\Return \( \{r_1, r_2, \ldots, r_i\} \)\ to participants;

\end{algorithm}

\begin{algorithm}[t]
      \For{$\forall e \in E$}
      {
      	 	\For{$\forall b \in B$}
       		{
       			$w_{i}\longleftarrow w_{i} - \eta \nabla \ell(w;b)$
       		}
       		Record $l_i^{e}$\\%, l_{i}^{CPU}, l_{i}^{RAM}$ \\
                %${Loss}_e = \mathcal{L}(y, \hat{y})$\\
       		%$TotalLoss=TotalLoss + Loss_{e}^{2}$
      }
      Calculate $H(L)_i, H(Y)_i, G_i$ \\
      Send $[H(L)_i, H(Y)_i, G_i]$ to the server \\
      Wait server calculate rank ratio $r_i$ \\
      Receive $r_i$ and initialize local Low-Rank model \\
      Start training
    \caption{Procedure executes on the participant. $H(L)_i, H(Y)_i, G_i$ represent the loss information entropy, distribution entropy and Gini-Simpson index respectively.}\label{alg:client}
\end{algorithm}\setlength{\textfloatsep}{2 mm}

\section{Experiment and Evaluation}\label{sec:experiment_and_evaluation}
In our study, we evaluate the effectiveness of AutoRank using MLP and CNN architectures on the MNIST, FMNIST, CIFAR-10\cite{krizhevsky2009learning}, and CINIC-10 \cite{darlow2018cinic} datasets. The based line methods we compared against are: \textbf{FedLoRA}\cite{yi2023fedlora} and \textbf{FlexLoRA}\cite{bai2024federated}, with the homogeneous rank; \textbf{RBLA} \cite{chen2024rbla}, with manual assigned heterogeneous rank. To ensure reproducibility, we conduct all experiments with a \textbf{fixed seed of 42} and the same configuration.

\subsection{Experiments setup}
\subsubsection{Baseline method}
To our best knowledge, currently there is no method designed for set initial rank for participants in distributed machine learning. So we pick FedLoRA with homogeneous rank, and RBLA with heterogeneous rank. { \textcolor{black}{Since SVD based LoRA aggregation methods will loss significant information to generate heterogeneous rank low rank model. We use the lossless method RBLA as our aggregation method to perform a more meaningful analysis.}
\subsubsection{Model configurations}
In the experiment, the MLP model for MNIST and FMNIST consists of two hidden layers, each with 200 neurons and ReLU activation, followed by a 10-class softmax output layer designed for flattened 28x28 pixel images (784-dimensional vectors). For CIFAR and CINIC datasets, we utilized a more complex architecture comprising three convolutional layers: batch normalization, max pooling, dropout, and fully connected layers. The first two convolutional layers include 32 and 64 filters, respectively, with a 3x3 kernel size and ReLU activation. These layers are followed by pooling layers for downsampling and dropout layers for regularization. The extracted feature maps are then flattened and passed through four fully connected layers, each with 512 neurons, interspersed with dropout layers. The final output is generated via a 10-class softmax layer. The optimizer for the CIFAR and CINIC experiments was Adam \cite{kingma2015adam}, and LoRA was applied exclusively to the dense layers in all experiments.

\subsection{non-IID \& Experiment Settings}

\begin{figure}[t]
  \centering
    \includegraphics[height = 42.5 mm]{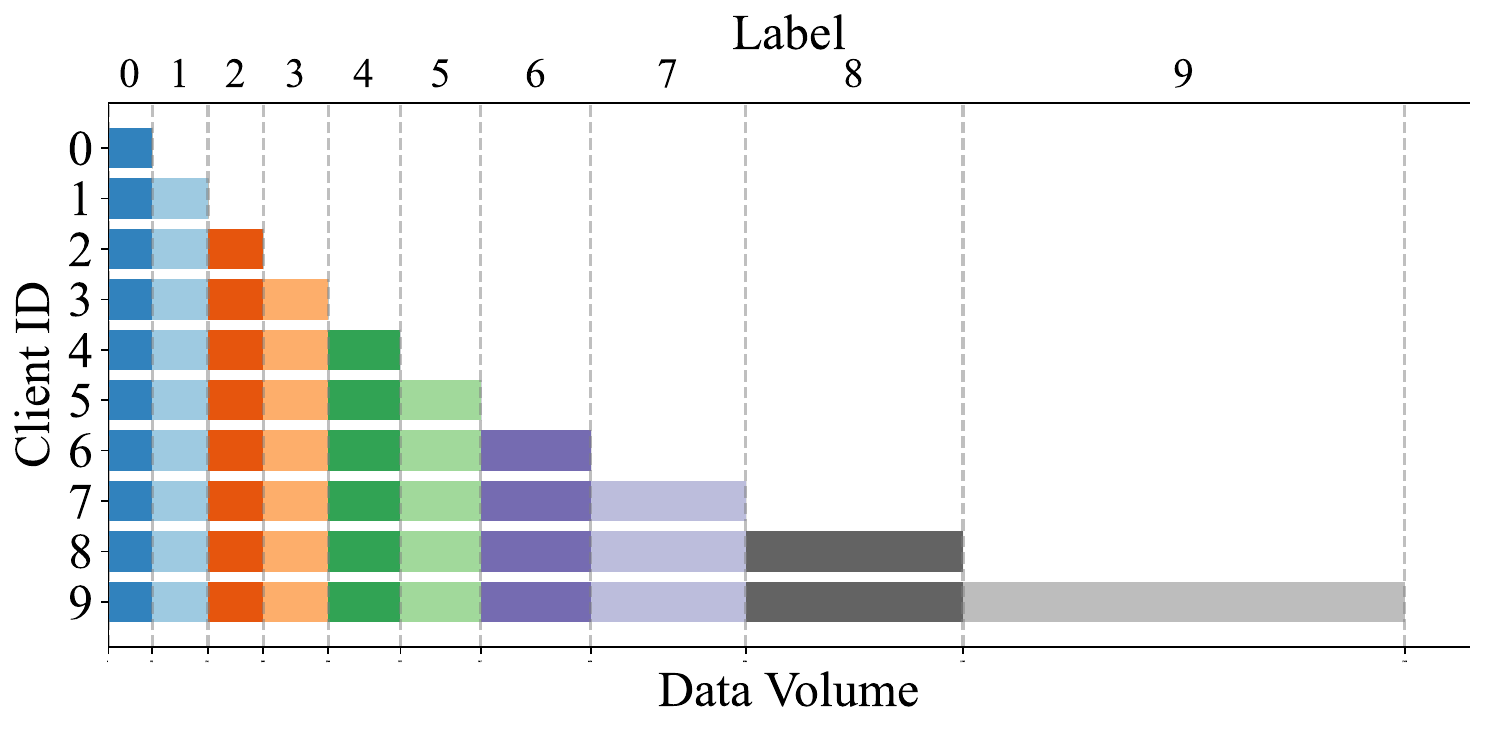}
  \caption{Data distribution in each participant in our experiment, we can see the client 9 owns all labels and much more data samples compared with other clients.} 
  \label{fig:longtail_distribution_ours}
\end{figure}

\subsubsection{Double imbalance non-IID settings}
In our experiments, data is distributed among participants following a distinct double imbalance \cite{6170916} "stair-case" patterns shown in Fig. \ref{fig:longtail_distribution_ours} to reflect real-world scenarios \textbf{where severe class imbalances within each individual task (imbalance in both data samples and label categories)} such as those found in medical and IoT systems. For example, data from different hospitals or clinics often vary significantly in complexity, quantity, and diversity due to their specialized roles and regional differences. Specialized hospitals, such as oncology or cardiology centers, focus primarily on specific types of diseases, resulting in a concentrated dataset with a limited number of labels but potentially complex and high-quality samples. In contrast, general hospitals encounter a broader spectrum of conditions and typically handle a larger volume of patients, leading to a more diverse dataset with a wider range of labels, albeit with potential variability in data consistency. Similarly, IoT sensing device networks can be deployed in environments with differing operational requirements and data generation patterns. For instance, sensors in industrial settings might prioritize monitoring specific machinery, leading to highly focused data streams. In contrast, sensors in smart city environments collect a diverse range of information, such as temperature, air quality, and traffic patterns. This disparity can result in non-uniform distributed data and heterogeneous labels across devices. To simulate this diversity, we allocate an increasing number of labels with non-zero sample counts to each subsequent participant. Participant 1, for instance, contains samples only for Label 0. As more participants are added, they are progressively assigned additional labels, culminating in participant 10, which has a significant number of samples that span all labels from 0 to 9. Fig.~\ref{fig:local_test_accuracy} compares the test accuracy of each participant's local model on the global test set accuracy compared to the global model. While participant 9 achieves significantly higher test accuracy than other participants, indicating its local data is highly representative, the data from participants 0 to 8 provide critical supplementary information to the global model. Through rank setting enabled by the AutoRank method, the collaborative contributions of all participants substantially improve the overall performance of the global model (denoted as G). This highlights that, although some participants may perform exceptionally well, the diversity and contributions of all participants are essential for building a stronger global model in distributed learning.

\begin{figure}[!t]
    \centering
    % \begin{subfigure}[b]{0.24\textwidth}
    %     \includegraphics[height=42.5 mm]{paper_figures/experiment_analyze/mnist_local_test_accuracy.pdf}
    %     \caption{MNIST.}
    %     \label{fig:mnist_local_test_accuracy}
    % \end{subfigure}
    \begin{subfigure}[b]{0.24\textwidth}
        \includegraphics[height=42.5 mm]{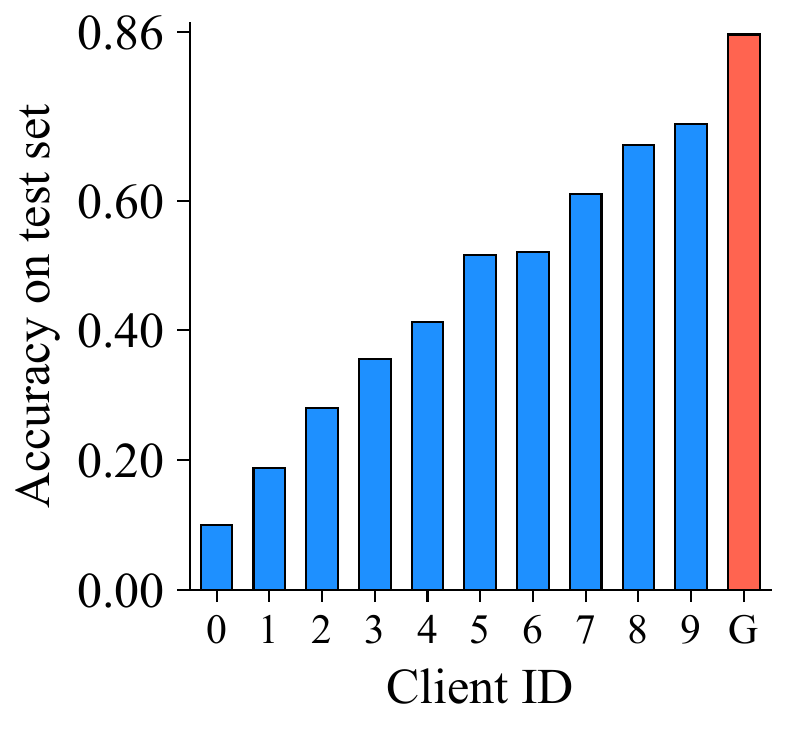}
        \caption{FMNIST.}
        \label{fig:fmnist_local_test_accuracy}
    \end{subfigure}
    % \begin{subfigure}[b]{0.24\textwidth}
    %     \includegraphics[height=42.5 mm]{paper_figures/experiment_analyse/cifar_local_test_accuracy.pdf}
    %     \caption{CIFAR}
    %     \label{fig:cifar_local_test_accuracy}
    % \end{subfigure}
    \begin{subfigure}[b]{0.24\textwidth}
        \includegraphics[height=42.5 mm]{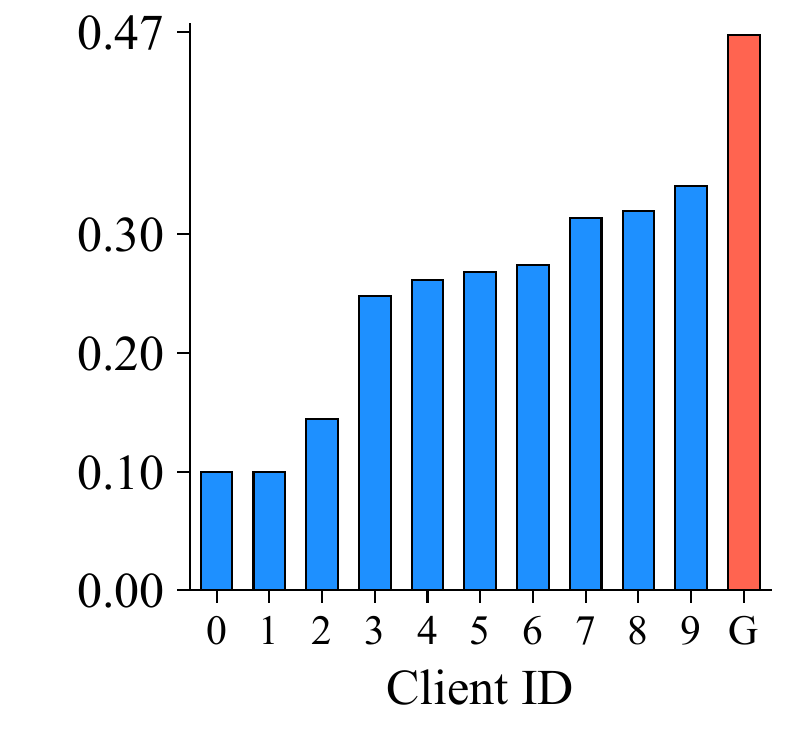}
        \caption{CINIC.}
        \label{fig:cinic_local_test_accuracy}
    \end{subfigure}
    \caption{Test accuracy on the global test set for each client's local model and the global model. Though clients (e.g., Client 9) achieve relatively high performance due to diverse and larger data distributions, their accuracy is still significantly lower than the global model $G$.}%\vspace{-4 mm}
    \label{fig:local_test_accuracy}
\end{figure}

\subsubsection{Heterogeneous rank settings}
The rank ratio of the LoRA model assigned to each participant is scaled proportionally to the number of labels the participant possesses. Specifically, the rank ratio increases by 0.1 for each additional label. \textbf{This manual allocation strategy is designed to be smooth and intuitive, which can be considered an ideal manual assigned rank from human prospective.} The ranks ensure that participants with more labels, more complex data distributions are assigned higher ranks to effectively capture their data complexity. Meanwhile, participants with fewer labels are assigned lower ranks, aligning the model's capacity with the diversity of their data. Such a balanced and gradual adjustment contributes to smoother convergence, making this an ideal scenario for manual rank assignment.

\begin{table}[b]
\caption{The highest test accuracy of each method can reach in different dataset.}
\centering
\resizebox{0.5\textwidth}{!}{%
\begin{tabular}{ccccccc}
\toprule
& MNIST & FMNIST & CIFAR & CINIC\\
Method & MLP & MLP & CNN & CNN \\
\midrule
RBLA  & 94.53\% & 84.79\% & 54.61\% & 42.76\%\\
FedLoRA  & 95.87\% & 85.03\% & 53.02\% & 45.60\% \\
FlexLoRA  & 95.71\% & 78.92\% & 31.01\% & 20.28\% \\
\midrule
Ours &\textbf{96.15\%} & \textbf{85.68\%} & \textbf{56.2\%}& \textbf{47.10\%} \\
\bottomrule
\end{tabular}}%\vspace{-2mm}
%\caption{The communication rounds each method takes to reach the highest test accuracy.}
\label{tab:result}
\end{table}

\subsection{Evaluation results}

\begin{figure*}[t]
    \centering
    \begin{subfigure}[b]{0.24\textwidth}
        \includegraphics[width=\textwidth]{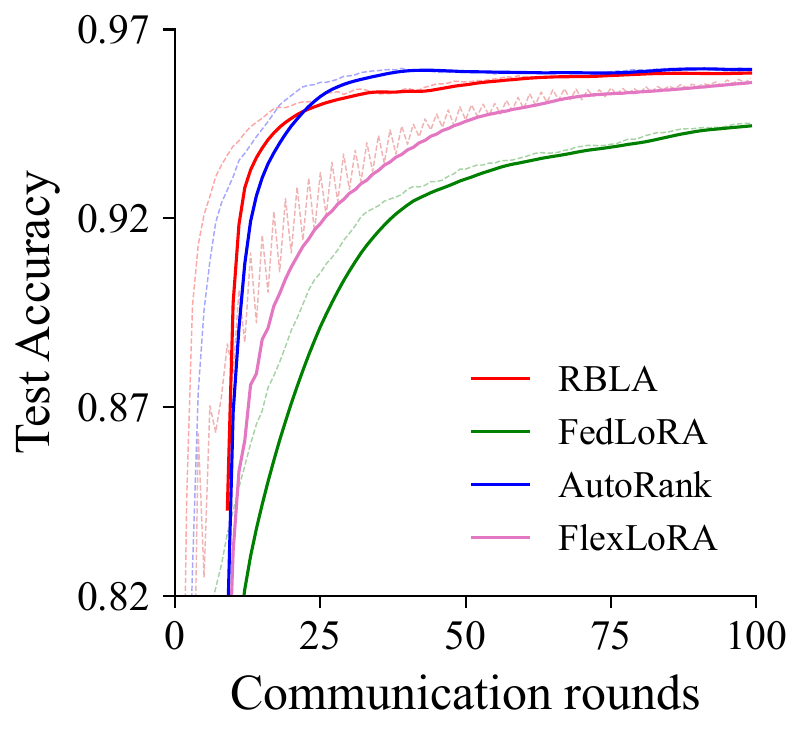}
        \caption{MNIST results.}
        \label{fig:mnist_all_participate}
    \end{subfigure}
    \hspace{-1.5em}
    \begin{subfigure}[b]{0.24\textwidth}
        \includegraphics[width=\textwidth]{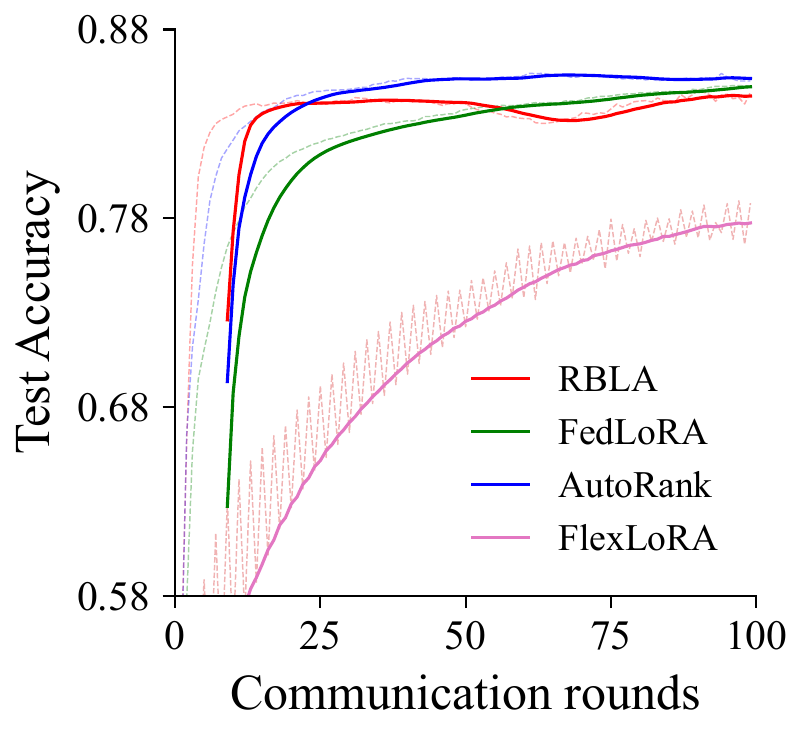}
        \caption{FMNIST results.}
        \label{fig:fmnist_all_participate}
    \end{subfigure}
    \hspace{-1.5em}
    \begin{subfigure}[b]{0.24\textwidth}
        \includegraphics[width=\textwidth]{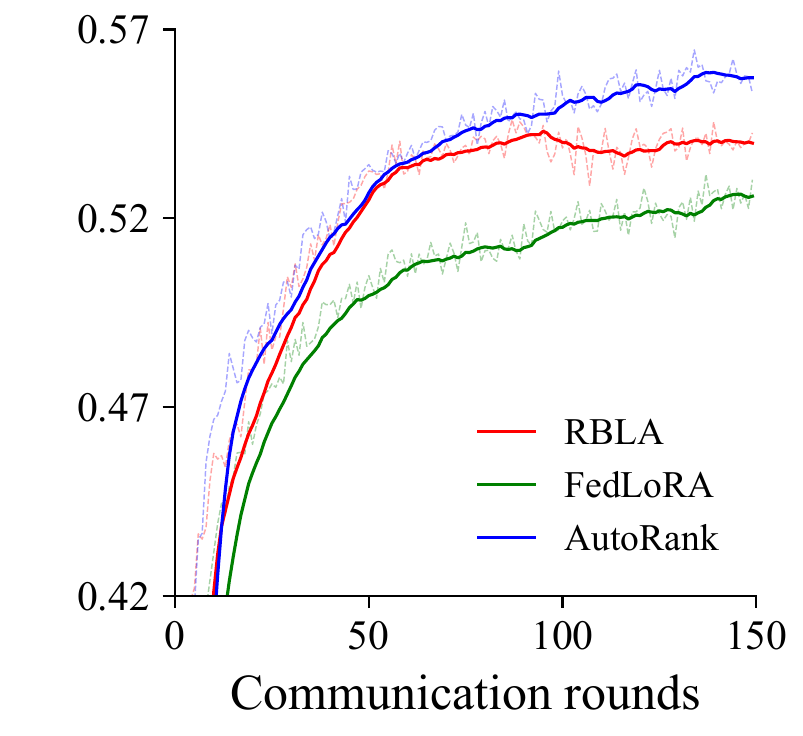}
        \caption{CIFAR-10 results.}
        \label{fig:cifar_all_participate}
    \end{subfigure}
    \hspace{-1.5em}
    \begin{subfigure}[b]{0.24\textwidth}
        \includegraphics[width=\textwidth]{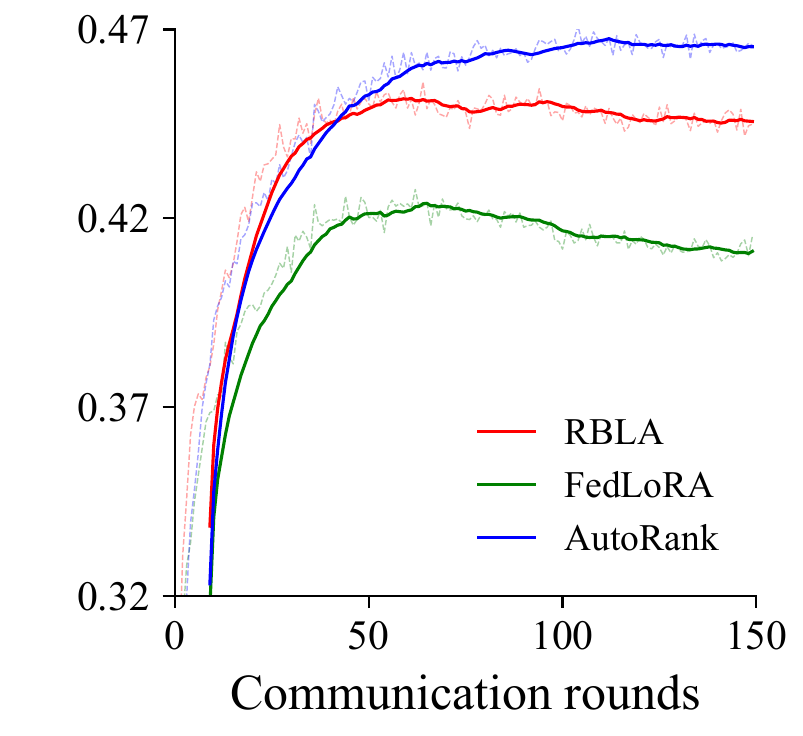}
        \caption{CINIC-10 results.}
        \label{fig:cinic_all_participate}
    \end{subfigure}
    \caption{Learning curve comparison in different datasets and rank setting methods.}
    \label{fig:experiments}
\end{figure*}

\begin{figure*}[t]
    \centering
    \begin{subfigure}[b]{0.27\textwidth}
        \includegraphics[width=\textwidth]{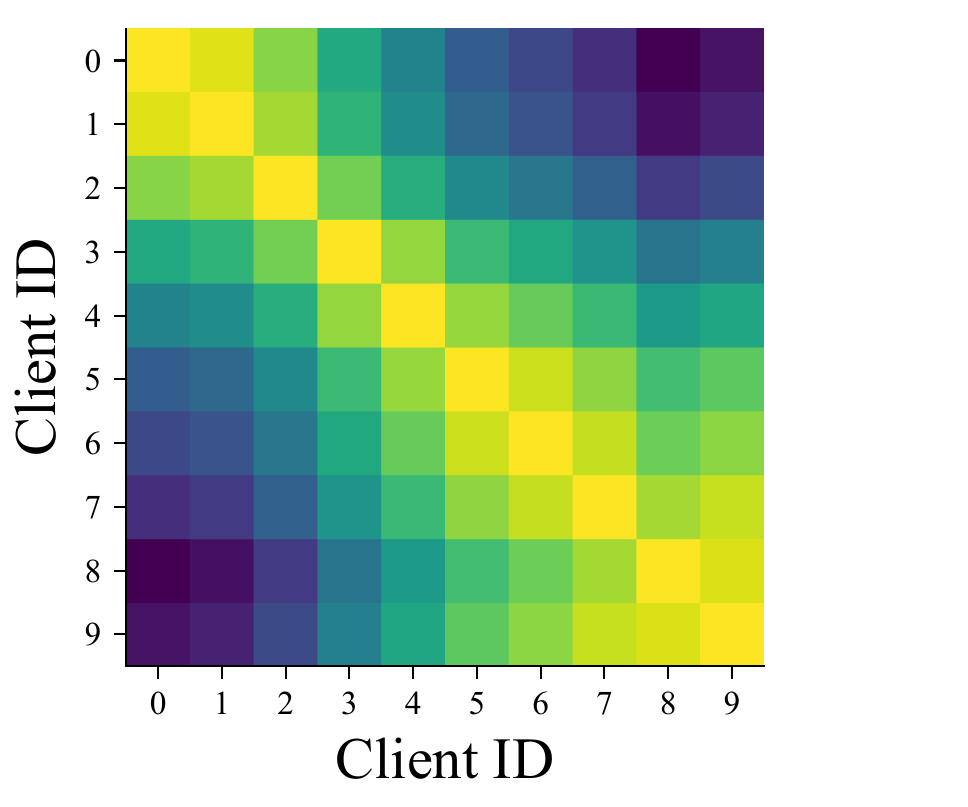}
        \caption{MNIST.}
        \label{fig:mnist_similarity}
    \end{subfigure}
    \hspace{-3.5em}
    \begin{subfigure}[b]{0.27\textwidth}
        \includegraphics[width=\textwidth]{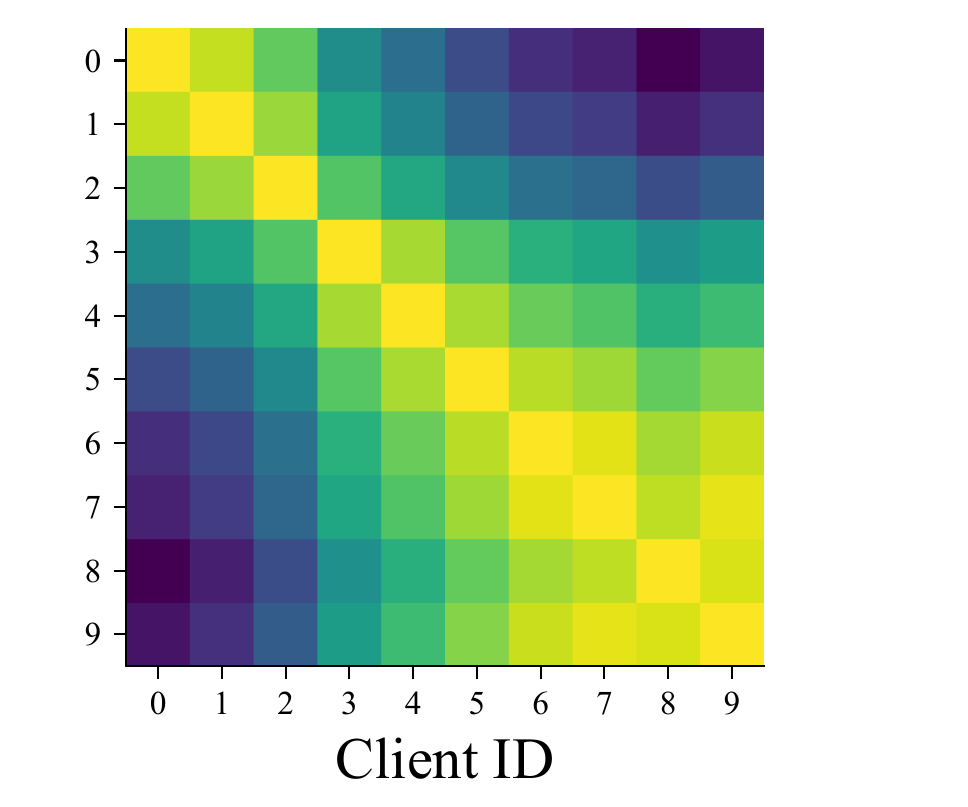}
        \caption{FMNIST.}
        \label{fig:fmnist_similarity}
    \end{subfigure}
    \hspace{-3.5em}
    \begin{subfigure}[b]{0.27\textwidth}
        \includegraphics[width=\textwidth]{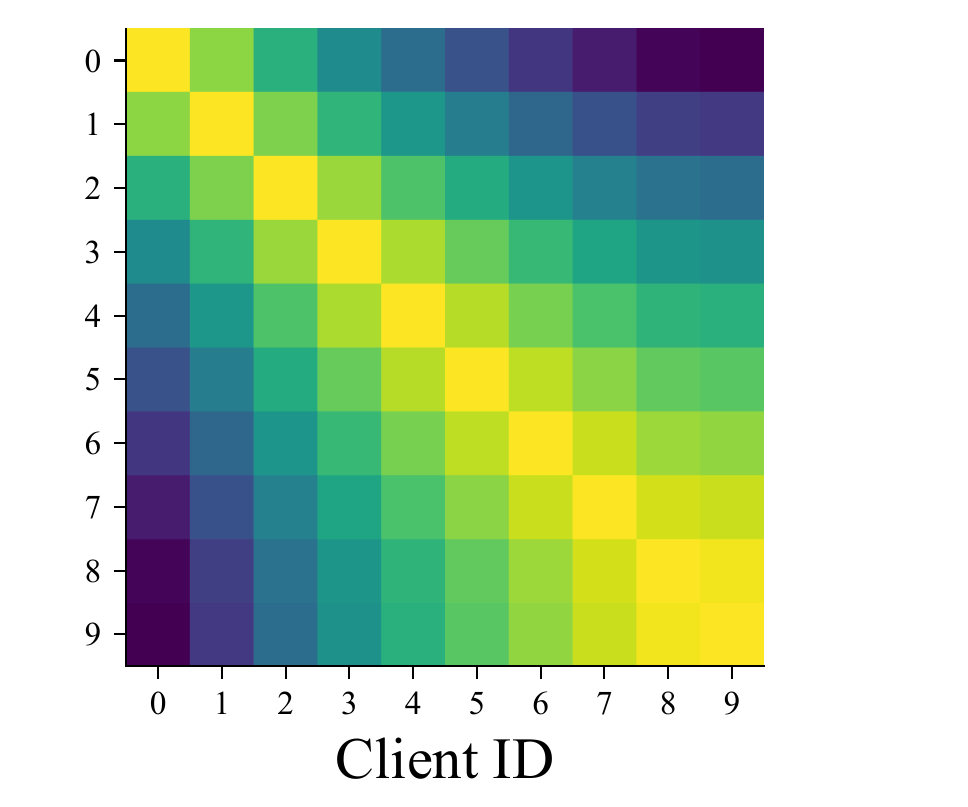}
        \caption{CIFAR-10.}
        \label{fig:cifar_similarity}
    \end{subfigure}
    \hspace{-3.5em}
    \begin{subfigure}[b]{0.27\textwidth}
        \includegraphics[width=\textwidth]{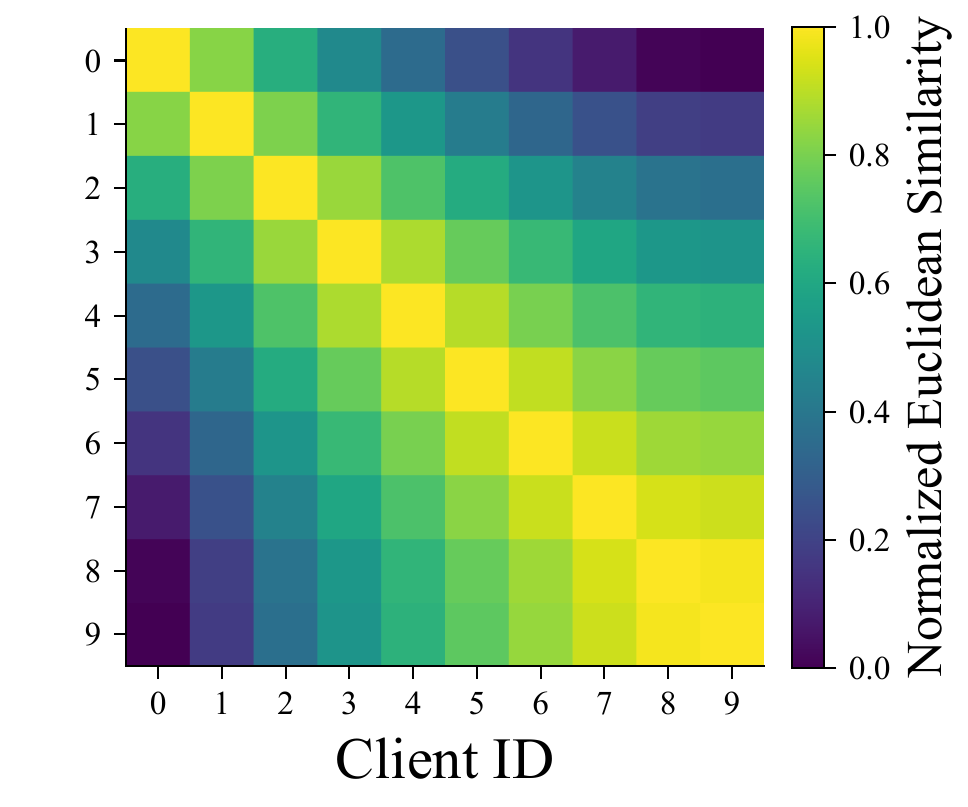}
        \caption{CINIC-10.}
        \label{fig:cinic_similarity}
    \end{subfigure}
    \caption{Similarity comparison of participant rank generated by AutoRank. The lighter the color, the lower the rank is assigned.}
    \label{fig:similarity_comparison}
\end{figure*}

In this section, we present the evaluation results across various datasets with different configurations. Table~\ref{tab:result} shows the exact number of communication rounds taken for each method to reach the highest test accuracy of the global model. Fig.~\ref{fig:mnist_all_participate} to Fig.~\ref{fig:cinic_all_participate} show the learning curve of the global model's test accuracy to training rounds, which aim to present the model convergence rate under different methods; \textbf{Since accuracy fluctuation are common across figures, to ensure clear visual comparisons, a rolling average with a window size of 10 was applied to smooth the data}, represented by solid lines, and the dotted lines illustrate the original and unsmoothed results. Fig.~\ref{fig:mnist_similarity} to Fig.~\ref{fig:cinic_similarity} illustrate the visualization of rank similarity among all participants. The color intensity represents the normalized Euclidean similarity between participant ranks, where brighter colors indicate higher similarity. The effectiveness of AutoRank is assessed on the MNIST, FMNIST, CIFAR-10, and CINIC-10 datasets using MLP and CNN models.

Fig.~\ref{fig:experiments} and Fig.~\ref{fig:similarity_comparison} provide a comprehensive evaluation of AutoRank's performance and robustness. Fig. 7 showcases the learning curve comparisons for three methods—RBLA, FedLoRA, and AutoRank. Specifically, AutoRank achieves the fastest convergence and highest test accuracy across all datasets. In simpler datasets like MNIST and FMNIST, AutoRank quickly outpaces RBLA, FedLoRA, and FlexLoRA, demonstrating its strong adaptability to heterogeneous data. On more complex datasets such as CIFAR-10 and CINIC-10, AutoRank continues to outperform, maintaining robust performance despite increased data heterogeneity, where RBLA achieves moderate results, and FedLoRA shows significant limitations. Fig.~\ref{fig:similarity_comparison} complements these results by visualizing the normalized Euclidean similarity of participant ranks generated by AutoRank. The more uniform the color, the more discrete the rank distributes. The rank distribution of MNIST and FMNIST is more discrete because they do not need too many ranks to capture the data of simple clients. For CIFAR and CINIC, due to the increased data complexity, larger models are needed to better fit local data, so AutoRank assigns higher ranks to clients with simple data. Which demonstrates that \textbf{AutoRank can dynamically adjust the rank distribution according to the complexity of the learning task}. Together, Fig.~\ref{fig:experiments} and Fig.~\ref{fig:similarity_comparison} demonstrate AutoRank's capability and robustness to adapt dynamically under different learning difficulty tasks.

\begin{figure}[t]
  \centering
    \includegraphics[height = 42.5 mm]{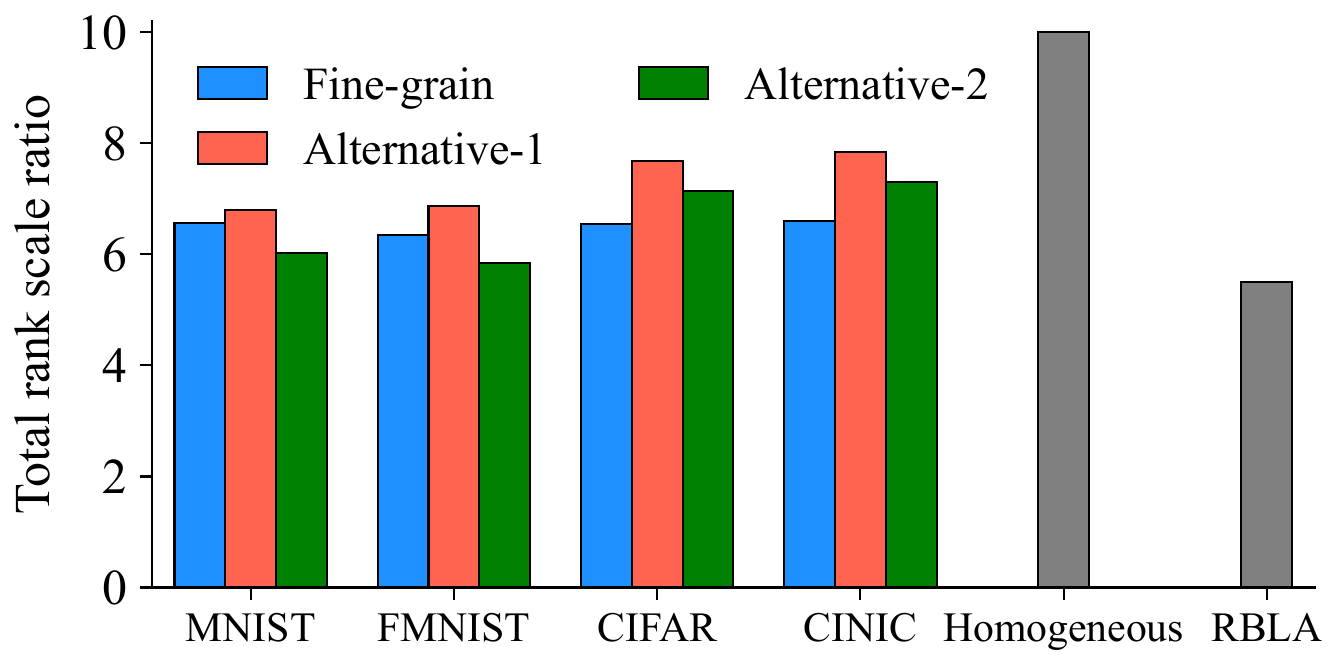}
  \caption{Sum of rank ratio from each participant under different configurations and datasets, the Homogeneous tick represents the rank ratio for FedLoRA and FlexLoRA when total rank ratio equal to 10, the overall trainable parameter for CNN in all participants is 20.39M, for MLP is 1.99M.} 
  \label{fig:total_rank_scale_ratio}
\end{figure}

\begin{table}[b]
\caption{Metrics selection for different alternative AutoRank configurations. Both alternative solutions can be applied to scenarios where data are not labeled.}
\centering
\begin{tabular}{llll}
\hline
\textbf{Metric}             & \textbf{Fine-grain} & \textbf{Alternative-1} & \textbf{Alternative-2} \\ \hline
Loss Entropy (LE)                & $\times$     & $\times$             & \checkmark             \\
Distribution Entropy               & \checkmark          &  $\times$                      &   $\times$                     \\
Gini-Simpson Index          & \checkmark          &   $\times$                     &   $\times$                     \\
Log Data Volume (LDV)                 & $\times$          &   $\times$                    & \checkmark             \\
LE * LDV                 & \checkmark          &   \checkmark                    & $\times$     \\
\hline
\end{tabular}
% \caption{Metrics selection for different alternative AutoRank configurations. Both alternative solution can be applied to scenario where data are not labeled.}
\label{table:alternative_config}
\end{table}

\subsection{Flexibility Study}
Our method is highly flexible and can be adapted seamlessly to different scenarios by modifying the input parameters or evaluation metrics. In this section, we evaluate the adaptability of AutoRank in various configurations, highlighting its robustness and versatility. Table~\ref{table:alternative_config} summarizes the metric selections across these alternative configurations. We compare their performance against our fine-grained metric setting to demonstrate AutoRank's effectiveness.

Table~\ref{table:alternative_config} summarizes the metric choices for each approach. Fine-grain relies on label-dependent metrics (e.g., LE * LDV), restricting its usage to labeled (supervised) scenarios but offering more precise rank allocation. In contrast, Alternative-1 and Alternative-2 draw on label-free metrics such as Distribution Entropy and data volume, making them capable of unlabeled data scenarios. Fig.~\ref{fig:total_rank_scale_ratio} presents the rank scale ratios for different configurations. Taking data complexity into consideration, the fine-grain method allocates about 15\%-20\% more trainable parameters overall. Compared with Alternative-1 and Alternative-2, the allocation results are less affected by the difficulty of the learning task and are more robust.

% The Fine-grain method assigns ranks more stable compared with alternative solution. However, this added accuracy comes at the cost of higher complexity (from 115\% - 120\%). Meanwhile, Alternative-1 and Alternative-2 adopt simpler rank allocations but do not attain the same level of precision and accuracy. Although Fine-grain clearly outperforms FedLoRA and RBLA, its elevated complexity may limit its applicability in certain scenarios.

\begin{figure}[!t]
    \centering
    \begin{subfigure}[b]{0.24\textwidth}
        \includegraphics[width=45 mm, height=42.5 mm]{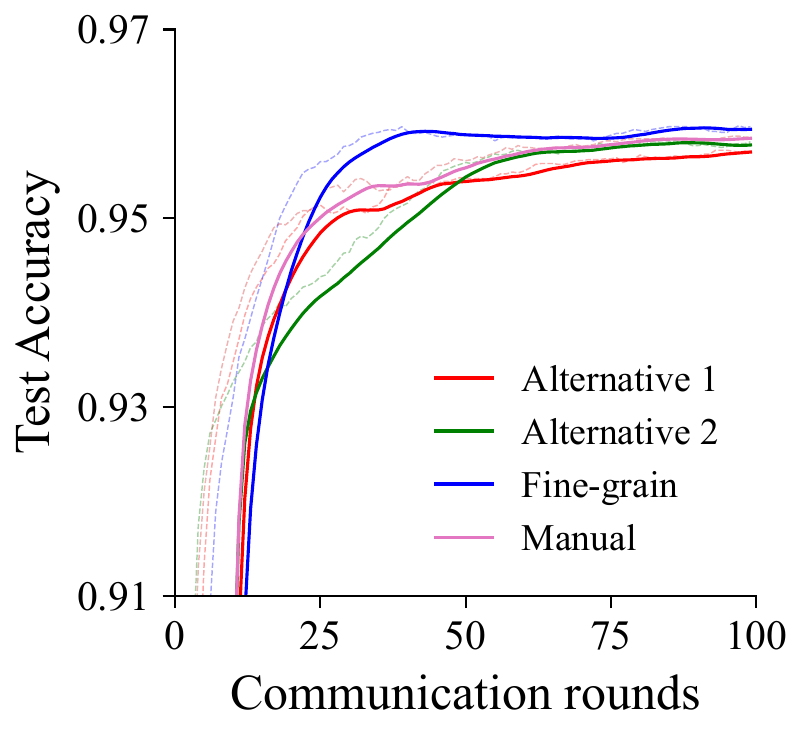}
        \caption{MNIST.}
        \label{fig:mnist_alternative}
    \end{subfigure}
    \begin{subfigure}[b]{0.24\textwidth}
        \includegraphics[width=45 mm, height=42.5 mm]{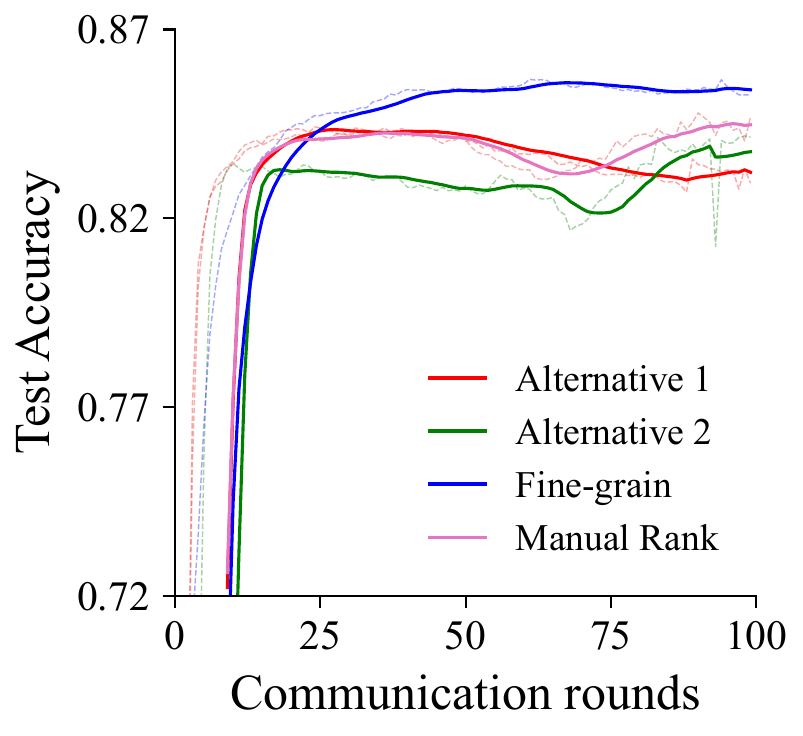}
        \caption{FMNIST.}
        \label{fig:fmnist_alternative}
    \end{subfigure}
\caption{Learning curve comparison AutoRank alternative solution by using \textbf{MNIST} and \textbf{FMNIST} dataset.}%\vspace{-4 mm}
\label{fig:alternative_learning_curve}
\end{figure}

\begin{figure}[!t]
    \centering
    \begin{subfigure}[b]{0.24\textwidth}
        \includegraphics[height=40 mm]{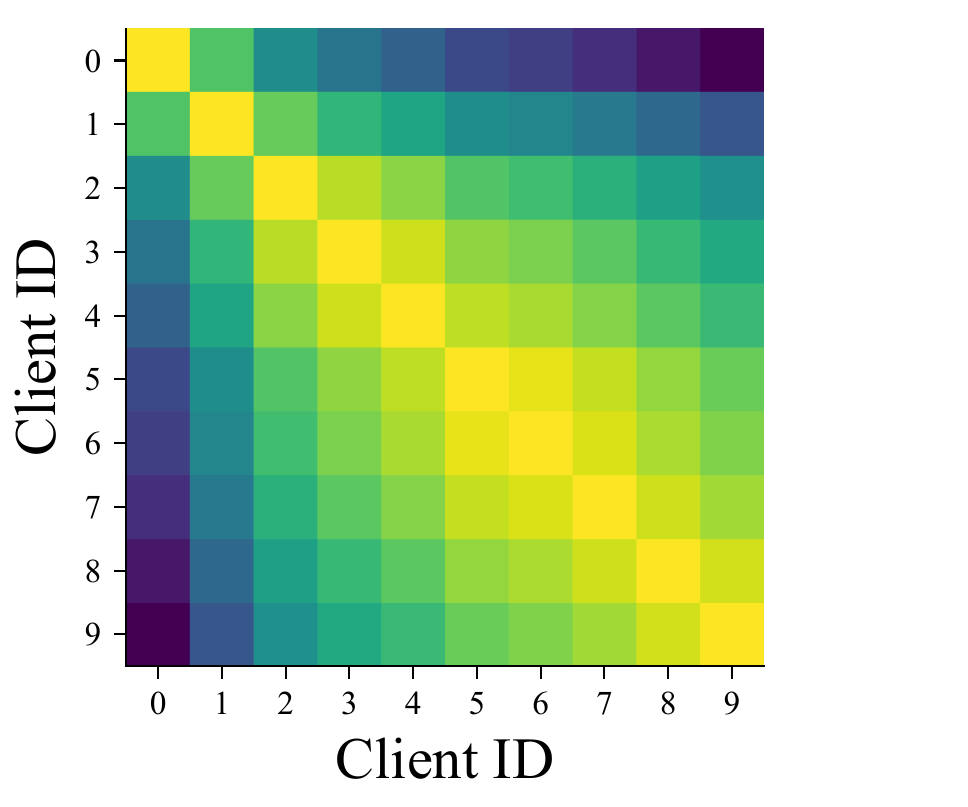}
        \caption{MNIST Alternative-1.}
        \label{fig:mnist_alternative_rank_similarity}
    \end{subfigure}
    \hspace{-1.5em}
    \begin{subfigure}[b]{0.24\textwidth}
        \includegraphics[height=40 mm]{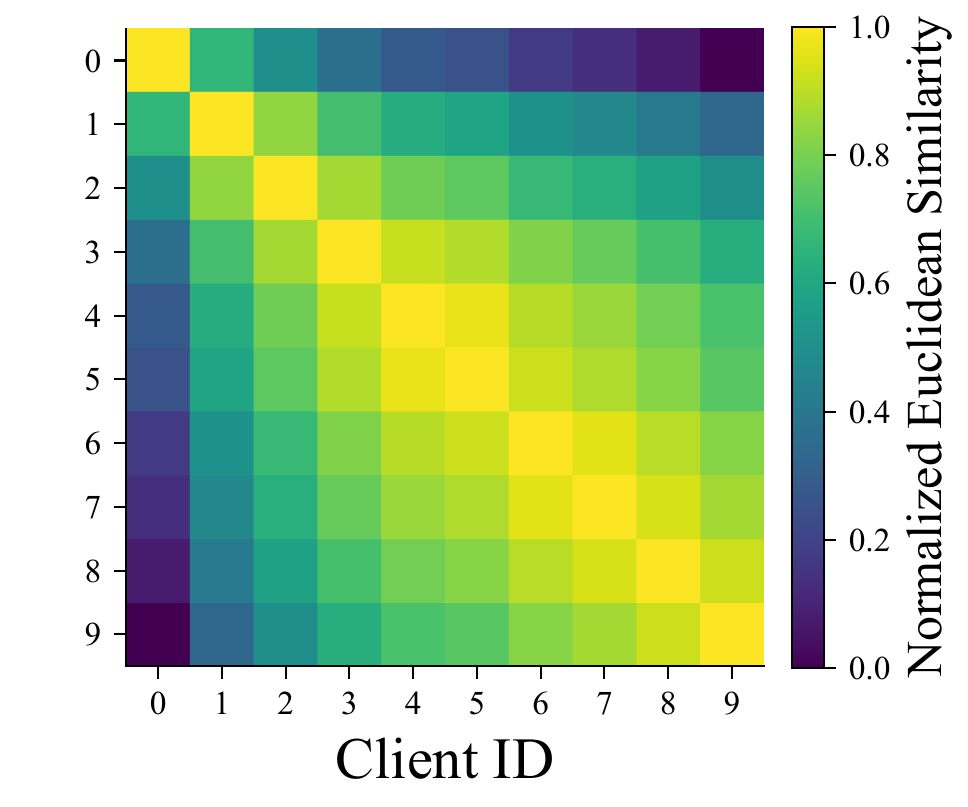}
        \caption{FMNIST Alternative-1.}
        \label{fig:fmnist_alternative_rank_similarity}
    \end{subfigure}

    \centering
    \begin{subfigure}[b]{0.24\textwidth}
        \includegraphics[height=40 mm]{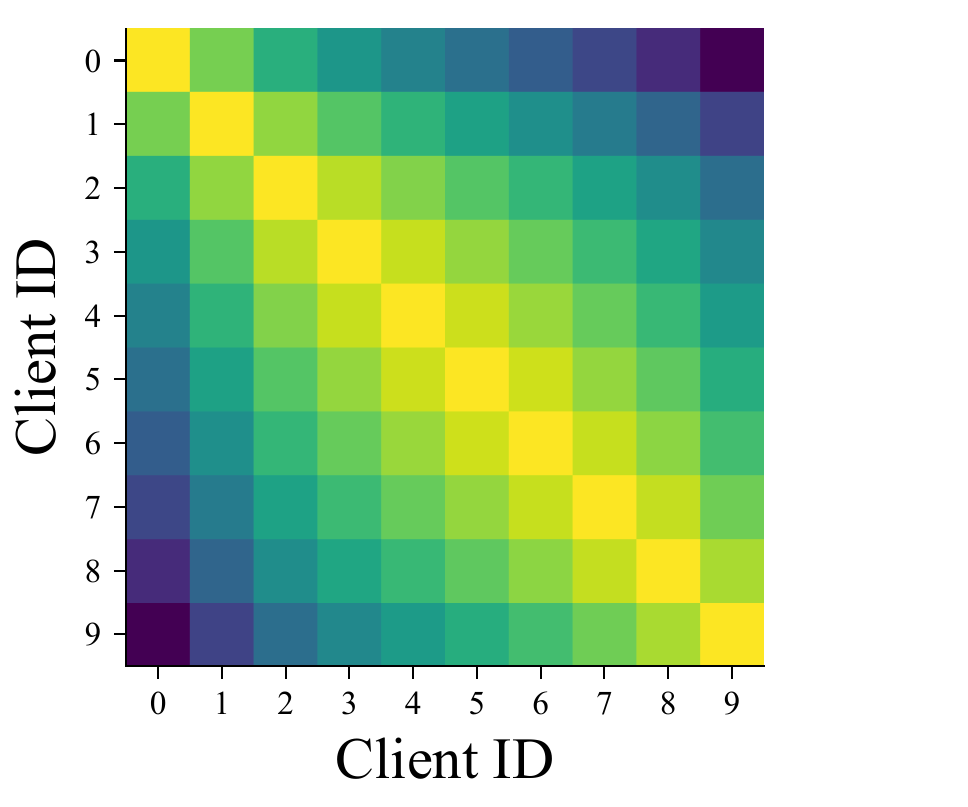}
        \caption{MNIST Alternative-2.}
        \label{fig:mnist_alternative_2_rank_similarity}
    \end{subfigure}
    \hspace{-1.5em}
    \begin{subfigure}[b]{0.24\textwidth}
        \includegraphics[height=40 mm]{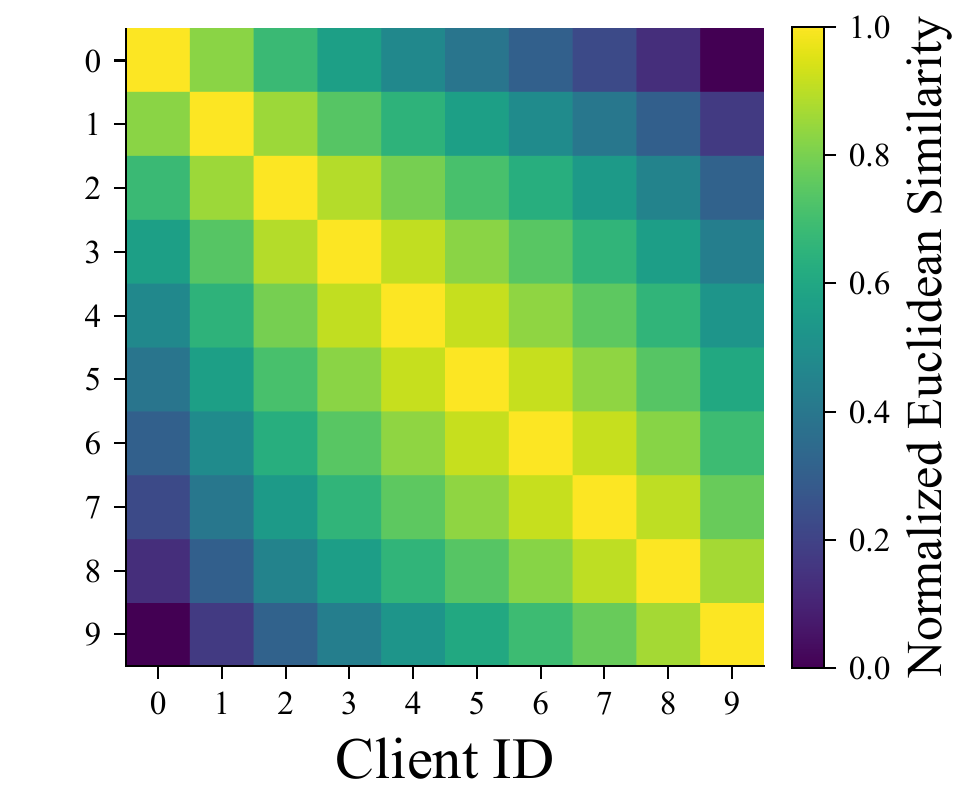}
        \caption{FMNIST Alternative-2.}
        \label{fig:fmnist_alternative_2_rank_similarity}
    \end{subfigure}
    
    \caption{Rank similarity analysis of different alternative solutions.}%\vspace{-4 mm}
    \label{fig:alternative_similarity}
\end{figure}
Fig.~\ref{fig:alternative_learning_curve} presents the learning curves for various configurations. On the MNIST and FMNIST datasets, the Fine-grain configuration converges the fastest and attains the highest final test accuracy, illustrating its ability to fully exploit data characteristics. Alternative-2 slightly surpasses Alternative-1 in test accuracy, indicating greater robustness and adaptability—yet both Alternatives exhibit comparable results overall, underscoring their suitability for resource-constrained or unlabeled data settings. Fig.~\ref{fig:alternative_similarity} visualizes the assigned rank similarity among participants. For Alternative-1 and Alternative-2, the heat maps reveal relatively less sparsity in assigned ranks than the Fine-grain solution. \textbf{Although the performance of the Alternative method is not as good as Fine-grain in our experiment, when it is impossible to use labels to estimate data complexity, the Alternative solution can still show results similar to manual allocation, greatly expanding its adaptability.}
In summary, AutoRank offers flexible rank allocation strategies. The Fine-grain method proves ideal for scenarios demanding high accuracy, leveraging labeled data to achieve optimal performance. By contrast, Alternative-1 and Alternative-2 emphasize balanced rank distribution, rendering them better suited for resource-limited or unlabeled tasks. Across diverse federated learning scenarios, AutoRank can be configured to achieve a robust equilibrium of performance, complexity, and applicability, serving as a versatile solution to meet varying requirements. %Moreover, AutoRank can serve as a pre-method for other LoRA-based distributed learning approaches and can be flexibly integrated into any distributed machine learning framework that adopts LoRA techniques.

\section{Conclusion}\label{section:conclusion}
In this paper, drawing on the bias-variance tradeoff, we conducted a comprehensive analysis of how local model complexity interacts with the global model’s generalization ability in distributed machine learning. Building on these insights, we proposed an evaluation framework to quantify local model complexity. Leveraging that framework, we further introduced AutoRank, an automated method for setting the LoRA rank in distributed environments. And, we evaluated the effectiveness of AutoRank under practical double imbalance non-IID scenarios and analyzed the flexibility of our method. The experiments demonstrated the practicality and efficiency of AutoRank, showing its adaptability for real-world distributed learning tasks. In the future, AutoRank can be extended to accommodate additional metrics for adaptive rank personalization, further improving scalability and performance in reality.

\bibliography{citation.bib}

\begin{thebibliography}{10}
\providecommand{\url}[1]{#1}
\csname url@samestyle\endcsname
\providecommand{\newblock}{\relax}
\providecommand{\bibinfo}[2]{#2}
\providecommand{\BIBentrySTDinterwordspacing}{\spaceskip=0pt\relax}
\providecommand{\BIBentryALTinterwordstretchfactor}{4}
\providecommand{\BIBentryALTinterwordspacing}{\spaceskip=\fontdimen2\font plus
\BIBentryALTinterwordstretchfactor\fontdimen3\font minus \fontdimen4\font\relax}
\providecommand{\BIBforeignlanguage}[2]{{%
\expandafter\ifx\csname l@#1\endcsname\relax
\typeout{** WARNING: IEEEtran.bst: No hyphenation pattern has been}%
\typeout{** loaded for the language `#1'. Using the pattern for}%
\typeout{** the default language instead.}%
\else
\language=\csname l@#1\endcsname
\fi
#2}}
\providecommand{\BIBdecl}{\relax}
\BIBdecl

\bibitem{mcmahan2017communication}
B.~McMahan, E.~Moore, D.~Ramage, S.~Hampson, and B.~A.~y. Arcas, ``{Communication-Efficient Learning of Deep Networks from Decentralized Data},'' in \emph{Proceedings of the 20th International Conference on Artificial Intelligence and Statistics}, vol.~54.\hskip 1em plus 0.5em minus 0.4em\relax PMLR, 2017, pp. 1273--1282.

\bibitem{9052677}
S.~Deng, H.~Zhao, W.~Fang, J.~Yin, S.~Dustdar, and A.~Y. Zomaya, ``Edge intelligence: The confluence of edge computing and artificial intelligence,'' \emph{IEEE Internet of Things Journal}, vol.~7, no.~8, pp. 7457--7469, 2020.

\bibitem{gao2024federated}
W.~Gao, O.~Tavallaie, S.~Chen, and A.~Zomaya, ``Federated learning as a service for hierarchical edge networks with heterogeneous models,'' in \emph{International Conference on Service-Oriented Computing}.\hskip 1em plus 0.5em minus 0.4em\relax Springer, 2024, pp. 85--99.

\bibitem{li2014scaling}
M.~Li, D.~G. Andersen, J.~W. Park, A.~J. Smola, A.~Ahmed, V.~Josifovski, J.~Long, E.~J. Shekita, and B.-Y. Su, ``Scaling distributed machine learning with the parameter server,'' in \emph{11th USENIX Symposium on operating systems design and implementation (OSDI 14)}, 2014, pp. 583--598.

\bibitem{dean2012large}
J.~Dean, G.~Corrado, R.~Monga, K.~Chen, M.~Devin, M.~Mao, M.~Ranzato, A.~Senior, P.~Tucker, K.~Yang \emph{et~al.}, ``Large scale distributed deep networks,'' \emph{Advances in neural information processing systems}, vol.~25, 2012.

\bibitem{vepakomma2018split}
P.~Vepakomma, O.~Gupta, T.~Swedish, and R.~Raskar, ``Split learning for health: Distributed deep learning without sharing raw patient data,'' in \emph{Proceedings of the ICLR AI for Social Good Workshop}, 2019.

\bibitem{10647104}
N.~Nazemi, O.~Tavallaie, S.~Chen, A.~Y. Zomaya, and R.~Holz, ``Boosting communication efficiency of federated learning’s secure aggregation,'' in \emph{2024 54th Annual IEEE/IFIP International Conference on Dependable Systems and Networks - Supplemental Volume (DSN-S)}, 2024, pp. 157--158.

\bibitem{shi2016edge}
W.~Shi, J.~Cao, Q.~Zhang, Y.~Li, and L.~Xu, ``Edge computing: Vision and challenges,'' \emph{IEEE internet of things journal}, vol.~3, no.~5, pp. 637--646, 2016.

\bibitem{han2015learning}
S.~Han, J.~Pool, J.~Tran, and W.~Dally, ``Learning both weights and connections for efficient neural network,'' \emph{Advances in neural information processing systems}, vol.~28, 2015.

\bibitem{krishnamoorthi2018quantizing}
R.~Krishnamoorthi, ``Quantizing deep convolutional networks for efficient inference: A whitepaper,'' \emph{arXiv preprint arXiv:1806.08342}, 2018.

\bibitem{hinton2015distilling}
G.~Hinton, ``Distilling the knowledge in a neural network,'' \emph{arXiv preprint arXiv:1503.02531}, 2015.

\bibitem{hu2021lora}
E.~J. Hu, Y.~Shen, P.~Wallis, Z.~Allen-Zhu, Y.~Li, S.~Wang, L.~Wang, and W.~Chen, ``Lora: Low-rank adaptation of large language models,'' \emph{arXiv preprint arXiv:2106.09685}, 2021.

\bibitem{howard2018universal}
J.~Howard and S.~Ruder, ``Universal language model fine-tuning for text classification,'' \emph{arXiv preprint arXiv:1801.06146}, 2018.

\bibitem{grativol2024flocora}
L.~Grativol, M.~Leonardon, G.~Muller, V.~Fresse, and M.~Arzel, ``Flocora: Federated learning compression with low-rank adaptation,'' in \emph{2024 32nd European Signal Processing Conference (EUSIPCO)}.\hskip 1em plus 0.5em minus 0.4em\relax IEEE, 2024, pp. 1786--1790.

\bibitem{xu2023qa}
Y.~Xu, L.~Xie, X.~Gu, X.~Chen, H.~Chang, H.~Zhang, Z.~Chen, X.~Zhang, and Q.~Tian, ``Qa-lora: Quantization-aware low-rank adaptation of large language models,'' \emph{arXiv preprint arXiv:2309.14717}, 2023.

\bibitem{dettmers2024qlora}
T.~Dettmers, A.~Pagnoni, A.~Holtzman, and L.~Zettlemoyer, ``Qlora: Efficient finetuning of quantized llms,'' \emph{Advances in Neural Information Processing Systems}, vol.~36, 2024.

\bibitem{wang2024flora}
Z.~Wang, Z.~Shen, Y.~He, G.~Sun, H.~Wang, L.~Lyu, and A.~Li, ``Flora: Federated fine-tuning large language models with heterogeneous low-rank adaptations,'' \emph{arXiv preprint arXiv:2409.05976}, 2024.

\bibitem{zhang2024towards}
J.~Zhang, S.~Vahidian, M.~Kuo, C.~Li, R.~Zhang, T.~Yu, G.~Wang, and Y.~Chen, ``Towards building the federatedgpt: Federated instruction tuning,'' in \emph{ICASSP 2024-2024 IEEE International Conference on Acoustics, Speech and Signal Processing (ICASSP)}.\hskip 1em plus 0.5em minus 0.4em\relax IEEE, 2024, pp. 6915--6919.

\bibitem{chen2024autoddl}
J.~Chen, S.~Li, R.~Guo, J.~Yuan, and T.~Hoefler, ``Autoddl: Automatic distributed deep learning with near-optimal bandwidth cost,'' \emph{IEEE Transactions on Parallel and Distributed Systems}, 2024.

\bibitem{yuan2021oneflow}
J.~Yuan, X.~Li, C.~Cheng, J.~Liu, R.~Guo, S.~Cai, C.~Yao, F.~Yang, X.~Yi, C.~Wu \emph{et~al.}, ``Oneflow: Redesign the distributed deep learning framework from scratch,'' \emph{arXiv preprint arXiv:2110.15032}, 2021.

\bibitem{li2020pytorch}
S.~Li, Y.~Zhao, R.~Varma, O.~Salpekar, P.~Noordhuis, T.~Li, A.~Paszke, J.~Smith, B.~Vaughan, P.~Damania \emph{et~al.}, ``Pytorch distributed: Experiences on accelerating data parallel training,'' \emph{arXiv preprint arXiv:2006.15704}, 2020.

\bibitem{kanpak2024cure}
H.~I. Kanpak, A.~Shabbir, E.~Gen{\c{c}}, A.~K{\"u}p{\c{c}}{\"u}, and S.~Sav, ``Cure: Privacy-preserving split learning done right,'' \emph{arXiv preprint arXiv:2407.08977}, 2024.

\bibitem{gentry2009fully}
C.~Gentry, ``Fully homomorphic encryption using ideal lattices,'' in \emph{Proceedings of the forty-first annual ACM symposium on Theory of computing}, 2009, pp. 169--178.

\bibitem{wainwright2005fundamental}
K.~Wainwright \emph{et~al.}, \emph{Fundamental methods of mathematical economics}.\hskip 1em plus 0.5em minus 0.4em\relax McGraw-Hill, 2005.

\bibitem{vapnik2015uniform}
V.~N. Vapnik and A.~Y. Chervonenkis, ``On the uniform convergence of relative frequencies of events to their probabilities,'' in \emph{Measures of complexity: festschrift for alexey chervonenkis}.\hskip 1em plus 0.5em minus 0.4em\relax Springer, 2015, pp. 11--30.

\bibitem{hastie2009elements}
T.~Hastie, ``The elements of statistical learning: data mining, inference, and prediction,'' 2009.

\bibitem{shwartz2017opening}
R.~Shwartz-Ziv and N.~Tishby, ``Opening the black box of deep neural networks via information,'' \emph{arXiv preprint arXiv:1703.00810}, 2017.

\bibitem{neyshabur2017exploring}
B.~Neyshabur, S.~Bhojanapalli, D.~McAllester, and N.~Srebro, ``Exploring generalization in deep learning,'' \emph{Advances in neural information processing systems}, vol.~30, 2017.

\bibitem{tzeng2011multiple}
G.-H. Tzeng and J.-J. Huang, \emph{Multiple attribute decision making: methods and applications}.\hskip 1em plus 0.5em minus 0.4em\relax CRC press, 2011.

\bibitem{10.1016/0305-0548(94)00059-H}
\BIBentryALTinterwordspacing
D.~Diakoulaki, G.~Mavrotas, and L.~Papayannakis, ``Determining objective weights in multiple criteria problems: the critic method,'' \emph{Comput. Oper. Res.}, vol.~22, no.~7, p. 763–770, Aug. 1995. [Online]. Available: \url{https://doi.org/10.1016/0305-0548(94)00059-H}
\BIBentrySTDinterwordspacing

\bibitem{krizhevsky2009learning}
A.~Krizhevsky and G.~Hinton, ``Learning multiple layers of features from tiny images,'' University of Toronto, Tech. Rep., 2009.

\bibitem{darlow2018cinic}
L.~N. Darlow, E.~J. Crowley, A.~Antoniou, and A.~J. Storkey, ``Cinic-10 is not imagenet or cifar-10,'' \emph{arXiv preprint arXiv:1810.03505}, 2018.

\bibitem{yi2023fedlora}
L.~Yi, H.~Yu, G.~Wang, and X.~Liu, ``Fedlora: Model-heterogeneous personalized federated learning with lora tuning,'' \emph{arXiv preprint arXiv:2310.13283}, 2023.

\bibitem{bai2024federated}
\BIBentryALTinterwordspacing
J.~Bai, D.~Chen, B.~Qian, L.~Yao, and Y.~Li, ``Federated fine-tuning of large language models under heterogeneous tasks and client resources,'' in \emph{The Thirty-eighth Annual Conference on Neural Information Processing Systems}, 2024. [Online]. Available: \url{https://openreview.net/forum?id=gkOzoHBXUw}
\BIBentrySTDinterwordspacing

\bibitem{chen2024rbla}
S.~Chen, O.~Tavallaie, N.~Nazemi, and A.~Y. Zomaya, ``Rbla: Rank-based-lora-aggregation for fine-tuning heterogeneous models in flaas,'' in \emph{International Conference on Web Services}.\hskip 1em plus 0.5em minus 0.4em\relax Springer, 2024, pp. 47--62.

\bibitem{kingma2015adam}
\BIBentryALTinterwordspacing
D.~P. Kingma and J.~Ba, ``Adam: A method for stochastic optimization,'' in \emph{3rd International Conference on Learning Representations (ICLR)}, 2015. [Online]. Available: \url{https://arxiv.org/abs/1412.6980}
\BIBentrySTDinterwordspacing

\bibitem{6170916}
S.~Wang and X.~Yao, ``Multiclass imbalance problems: Analysis and potential solutions,'' \emph{IEEE Transactions on Systems, Man, and Cybernetics, Part B (Cybernetics)}, vol.~42, no.~4, pp. 1119--1130, 2012.

\end{thebibliography}
\bibliographystyle{IEEEtran}
\end{document}